\documentclass{article}

\usepackage{xcolor}         %
\definecolor{darkblue}{RGB}{0,0,128}
\PassOptionsToPackage{colorlinks = True, linkcolor = darkblue, citecolor = darkblue}{hyperref}
\usepackage[accepted]{icml2024}
\usepackage[utf8]{inputenc} %
\usepackage[T1]{fontenc}    %
\usepackage[backref=page]{hyperref}       %
\usepackage{url}            %
\usepackage{booktabs}       %
\usepackage{amsfonts}       %
\usepackage{nicefrac}       %
\usepackage{microtype}      %
\usepackage{derivative}     %
\usepackage{stackengine}
\usepackage[export]{adjustbox}

\usepackage{subcaption}
\usepackage{mathtools}
\usepackage{amssymb}
\usepackage{amsmath}
\usepackage{wrapfig}
\usepackage{amsthm}

\newtheorem{prop}{Proposition}
\usepackage{multirow}

\usepackage{enumitem}
\usepackage{todonotes}
\usepackage{algorithm}
\usepackage{algorithmic}
\usepackage{placeins}
\renewcommand{\algorithmicrequire}{\textbf{Input:}}
\renewcommand{\algorithmicensure}{\textbf{Output:}}

\icmltitlerunning{Leveraging Self-Consistency for Data-Efficient Amortized Bayesian Inference}

\usepackage{amssymb}
\usepackage{amsmath}
\usepackage{bm}

\newcommand{\diff}{\mathrm{d}}
\newcommand{\given}{\,|\,}
\newcommand{\Var}{\operatorname{Var}}

\def\gL{\mathcal{L}}

\newcommand{\thetab}{\boldsymbol{\theta}}
\newcommand{\phib}{\boldsymbol{\phi}}

\newcommand{\etab}{\boldsymbol{\eta}}

\newcommand{\x}{\mathbf{x}}
\newcommand{\y}{\mathbf{y}}
\newcommand{\z}{\mathbf{z}}
\newcommand{\X}{\mathbf{X}}
\newcommand{\Y}{\mathbf{Y}}
\newcommand{\Z}{\mathbf{Z}}

\newcommand{\thetasc}{\thetab}

\begin{document}

\twocolumn[
\icmltitle{Leveraging Self-Consistency for Data-Efficient Amortized Bayesian Inference}

\begin{icmlauthorlist}
\icmlauthor{Marvin Schmitt}{1}
\icmlauthor{Desi R. Ivanova}{2}
\icmlauthor{Daniel Habermann}{3}
\icmlauthor{Ullrich Köthe}{4}
\icmlauthor{Paul-Christian Bürkner}{3}
\icmlauthor{Stefan T. Radev}{5}
\end{icmlauthorlist}

\icmlaffiliation{1}{University of Stuttgart, Germany}
\icmlaffiliation{2}{University of Oxford, UK}
\icmlaffiliation{3}{TU Dortmund University, Germany}
\icmlaffiliation{4}{Heidelberg University, Germany}
\icmlaffiliation{5}{Rensselaer Polytechnic Institute, USA}

\icmlcorrespondingauthor{Marvin Schmitt}{mail.marvinschmitt@gmail.com}

\icmlkeywords{Probabilistic Machine Learning, Simulation-Based Inference, Amortized Bayesian Inference}

\vskip 0.3in
]

\printAffiliationsAndNotice{}  %

\begin{abstract}
We propose a method to improve the efficiency and accuracy of amortized Bayesian inference by leveraging universal symmetries in the joint probabilistic model $p(\thetab, \Y)$ of parameters $\thetab$ and data $\Y$.
In a nutshell, we invert Bayes' theorem and estimate the marginal likelihood based on approximate representations of the joint model.
Upon perfect approximation, the marginal likelihood is constant across all parameter values by definition.
However, errors in approximate inference lead to undesirable variance in the marginal likelihood estimates across different parameter values.
We penalize violations of this symmetry with a \textit{self-consistency loss} which significantly improves the quality of approximate inference in low data regimes and can be used to augment the training of popular neural density estimators.
We apply our method to a number of synthetic problems and realistic scientific models, discovering notable advantages in the context of both neural posterior and likelihood approximation.
\end{abstract}

\vspace*{-2em}
\section{Introduction}
Computer simulations are ubiquitous in today's world, and their widespread application in the sciences has heralded a new era of \emph{simulation intelligence} \cite{lavin2021simulation}.
Typically, scientific simulators define a mapping from latent parameters $\thetab$ to observable data $\Y$.
This \emph{forward problem} is probabilistically described by the likelihood $p(\Y\given\thetab)$.
The \emph{inverse problem} of reasoning about the unknown parameters $\thetab$ given observed data $\Y$ and a prior $p(\thetab)$ is captured by the posterior $p(\thetab\given \Y) = p(\thetab)\,p(\Y\given\thetab) / p(\Y)$, 
which represents a coherent way to combine all available information in a probabilistic system \cite{BDA3} and quantify epistemic uncertainty \cite{hullermeier_aleatoric_2021}. 
For complex models, the marginal likelihood $p(\Y)$ is a high-dimensional integral, $p(\Y)=\int p(\thetab)\,p(\Y\given\thetab)\diff\thetab$, 
rendering the posterior analytically intractable in general.

In \emph{likelihood-based inference}, the likelihood is explicitly available as a probability density function $p(\Y\given\thetab)$, giving rise to a family of likelihood-based algorithms to approximate the posterior distribution.
Markov chain Monte Carlo (MCMC) methods sample from the unnormalized posterior by exploring the parameter space through a Markov chain, with state-of-the-art samplers such as Hamiltonian Monte Carlo \cite{neal_mcmc_2011}, as implemented in the probabilistic programming language Stan \cite{carpenter2017stan}.
Variational inference approximates the posterior via tractable analytic distributions, with consistent progress towards more trustworthy variational methods \cite{blei2017variational}.

Different from likelihood-based inference, \emph{simulation-based inference} (SBI) circumvents explicit likelihood evaluation and relies only on random samples from a simulation program $\Y \sim p(\Y, \Z \given \thetab)$ with latent program states or ``outsourced'' noise $\Z$ \cite{cranmer2020frontier}.
The execution paths of the simulation program define an implicit likelihood $p(\Y\given\thetab)=\int p(\Y,\Z\given\thetab)\diff \Z$, which is computationally intractable for any simulation program of practical interest.
However, we have access to samples $(\thetab, \Y)$ of parameter-data tuples by executing the simulation program repeatedly.
In the face of analytically intractable simulators, previous research has explored other properties of such programs for learning surrogate likelihood functions or the likelihood ratio \cite{brehmer2018guide, brehmer2020mining, brehmer2020madminer}.

As a general perspective on simulation intelligence, \emph{amortized Bayesian inference} (ABI) is concerned with enabling fully probabilistic SBI in real-time \cite{radev2020bayesflow, gonccalves2020training, avecilla2022neural}.
A core principle of ABI lies in tackling probabilistic problems (forward, inverse, or both) with neural networks.
By re-casting an intractable probabilistic problem as forward passes through a trained generative neural network, the required computational time reduces from hours (MCMC) to just a few seconds (ABI).
Yet, there is rarely a free lunch, and neural ABI algorithms require an upfront training phase. 
The associated effort is subsequently repaid with real-time inference on new data sets, thereby \emph{amortizing} the initial training time.

\begin{figure*}[t]
    \centering
    \begin{tabular}{c|c}
         \rotatebox{90}{\parbox{2.0cm}{\textbf{\;\;\;\;\;\;Perfect \mbox{\;\;\;\;Symmetry}}}}& \includegraphics[width=0.8\textwidth]{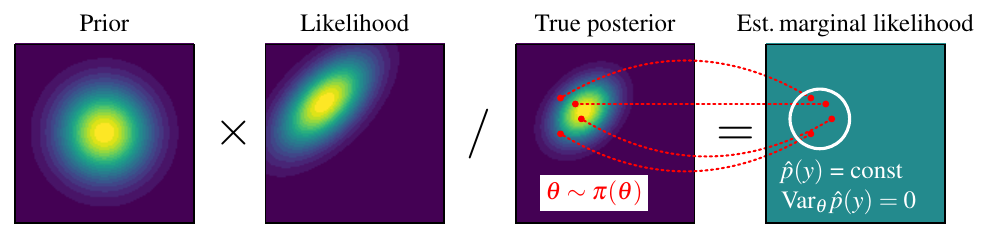}\\
         \hline
         \rotatebox{90}{\parbox{2.0cm}{\textbf{\;\;\;\;\;\;Posterior} \mbox{\textbf{\;Approximation}}}} & \includegraphics[width=0.8\textwidth]{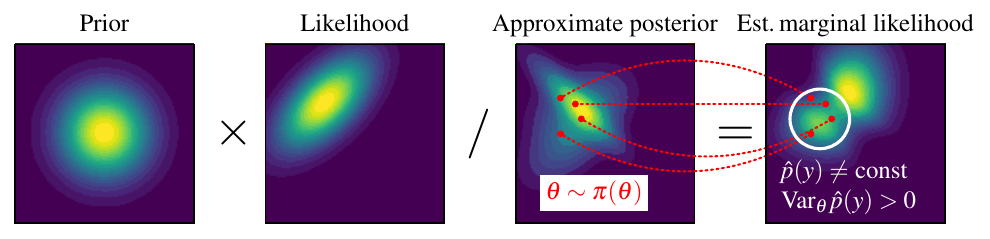} 
    \end{tabular}
    \caption{The performance of the posterior approximator is evaluated via the variance of the corresponding marginal likelihood estimates.
    \textbf{Top row:} For the true posterior (or a perfect approximation thereof), the estimated marginal likelihood %
    is constant for any parameter value $\smash{\thetasc\sim \pi(\thetab)}$.
    \textbf{Bottom row:} For an imperfect approximate posterior, the estimated marginal likelihood varies across different parameter values.
    Hence, the inherent symmetry of the joint probabilistic model $p(\thetab, \Y)$ is violated by its approximate representation.
    Minimizing the variance of the marginal likelihood estimates pushes the estimated marginal likelihood towards uniformity. 
    This restores the symmetry of the unified representation, which is equivalent to improving the approximate posterior.
    }
    \label{fig:main-figure}
\end{figure*}

In this paper, we propose a new method to improve the accuracy of ABI, particularly in low data regimes.
We achieve this by leveraging universal symmetries in the joint probabilistic model $p(\thetab, \Y)$.
In a nutshell, we invert Bayes' theorem and estimate the marginal likelihood based on an approximate likelihood and posterior.
Upon perfect approximation, the estimated marginal likelihood is constant across all parameters.
An imperfect approximation, however, leads to variance in the marginal likelihood estimates across different underlying parameter values (see \autoref{fig:main-figure}).
We penalize violations of this symmetry in the loss function to accelerate the training of neural density estimators that approximate components of the probabilistic model.
We apply our novel method to a range of synthetic and real problems with both an explicit likelihood (likelihood-based) and an implicit likelihood (simulation-based).
Our contributions are:
\begin{enumerate}[label=(\roman*), parsep=3pt, itemsep=3pt, topsep=3pt]
    \item We propose a novel \emph{self-consistency loss} which exploits symmetries of the joint probabilistic model $p(\thetab, \Y)$ in neural network representations of its components;
    \item We increase the training efficiency of amortized neural posterior estimation by leveraging information from an explicit likelihood for the self-consistency loss;
    \item We demonstrate how simultaneous learning of an approximate posterior and a surrogate likelihood benefits from the self-consistency loss without requiring an explicit likelihood.
\end{enumerate}

\section{Background}

\subsection{Notation}
Simulation-based training uses a \emph{training set} $\{(\thetab^{(i)}, \Y^{(i)})\}_{i=1}^N$ of tuples of simulated parameters and data.
Here, $N$ is the total \emph{simulation budget} for neural network training, and each superscript $i$ marks one training example (i.e., tuple of latent parameter vector and observable data set).
In accordance with the (Bayesian) forward model, we summarize the $D$-dimensional latent parameter vector of the simulation program as $\thetab\equiv(\theta_1, \ldots, \theta_D)$.
Further, $\Y\equiv\{\y_j\}_{j=1}^J$ is a data set, that is, a matrix whose rows consist of multi-dimensional (vector-valued) observations $\{\y_j\}_{j=1}^J\equiv\{\mathbf{y}_1, \ldots, \y_J\}$.
Accordingly, one parameter vector $\thetab^{(i)}$ yields one data set $\Y^{(i)}$.

\subsection{Neural Posterior Estimation}
Past work on neural SBI has primarily focused on neural posterior estimation (NPE) for cases where no explicit likelihood is available \cite{radev2020bayesflow, gonccalves2020training, avecilla2022neural,geffner2022diffusionsbi,sharrock2022diffusionsbi,schmitt2023cmpe}.
NPE typically builds on a conditional normalizing flow \citep{rezende2015} $f_{\phib}(\thetab;\Y)$ with trainable neural network weights $\phib$.
It implements a bijective map between the inference targets $\thetab\in\mathbb{R}^D$ and a latent variable $\z\in\mathbb{R}^D$ with a simple base distribution $p(\z)$, e.g., Gaussian or Student-$t$.
The normalizing flow defines a probability distribution $q_{\phib}(\thetab\given\Y)$ through change-of-variables,
\begin{equation}
    q_{\phib}(\thetab\given\Y) = p(\z=f_{\phib}(\thetab; \Y))\left\lvert\det\dfrac{\partial f_{\phib}(\thetab;\Y)}{\partial\thetab}\right\rvert,
\end{equation}
resulting in a direct surrogate for the target posterior distribution $p(\thetab\given\Y)$.
The neural network weights $\phib$ are trained by minimizing the forward Kullback-Leibler (KL) divergence between the target posterior and approximate posterior:
\begin{equation}
\begin{aligned}
    \mathcal{L}_{\text{NPE}}(\phib) &= \mathbb{E}_{p(\Y)}\left[\text{KL}(p(\thetab \given \Y) \, \Vert \, q_{\phib}(\thetab\given\Y))\right]\\ &= \mathbb{E}_{p(\thetab, \Y)}[-\log q_{\phib}(\thetab\given\Y)] + \text{const} \label{eq:loss_npe}
\end{aligned}
\end{equation}
Since NPE algorithms are trained across the entire prior predictive distribution $p(\Y)$, they naturally amortize over multiple new data sets during inference.
In \textbf{Experiments 1} and \textbf{4}, we apply NPE to cases where we \emph{have} access to an explicit likelihood but still want to achieve amortized inference, which is infeasible with MCMC-based algorithms.

\subsection{Neural Posterior and Likelihood Estimation}
Recently, simulation-based inference and surrogate modeling have been tackled \emph{jointly} by learning an approximate posterior $q_{\phib}(\thetab\given \Y)$ and a surrogate likelihood $q_{\etab}(\Y\given \thetab)$ in tandem.
This approach is called neural posterior and likelihood estimation \citep[NPLE;][]{wiqvist2021sequential,glockler2022snvi,radev2023jointly}.
Amortized NPLE \citep[e.g.,][]{radev2023jointly} combines the maximum likelihood objectives for posterior and likelihood into a joint loss:
\begin{equation}
    \mathcal{L}_{\text{NPLE}}(\phib, \etab) = \mathbb{E}_{p(\thetab,\Y)}[-\log q_{\phib}(\thetab\given\Y) - \log q_{\etab}(\Y\given\thetab)] \label{eq:loss_nple}
\end{equation}

NPLE enables the joint estimation of both amortized posterior predictive and marginal likelihood.
This expands the utility of Bayesian workflows to downstream tasks that have historically been deemed computationally impractical, such as cross-validation with likelihood-based predictive metrics \cite{radev2023jointly, vehtari2022loo}.

\subsection{Limitations of NPE and NPLE}
Whilst NPE and NPLE have shown promise in complex applications, they often require large simulation budgets and do not explicitly encourage accurate marginal likelihood estimation. 
In this paper, we propose a straightforward yet powerful approach to improve amortized Bayesian inference, integrating a self-consistency mechanism that alleviates these issues and enhances the accuracy of posterior and likelihood estimation with small simulation budgets.

\section{Leveraging Self-Consistency for ABI}\label{sec:method}

The joint model $p(\thetab, \Y)$ implies a symmetry between marginal likelihood $p(\Y)$, prior $p(\thetab)$, likelihood $p(\Y\given\thetab)$, and posterior $p(\thetab\given \Y)$. Inverting Bayes' theorem yields
\begin{equation}\label{eq:consistency}
p(\Y) = \dfrac{p(\thetab)\,p(\Y\given\thetab)}{p(\thetab\given \Y)},
\end{equation}
which must still hold if any component of the joint model is represented through a \emph{perfect} approximator $q(\cdot)$.
However, we cannot directly use Eq.~\ref{eq:consistency} as a loss function for learning the posterior because the marginal likelihood on the LHS is notoriously difficult to approximate with high precision \cite{meng_simulating_1996}.
Instead, we exploit the fact that $p(\Y)$ is constant across all parameters $\thetab$ (LHS), even though its computation (RHS) hinges on an arbitrary but fixed parameter value $\thetab$.
In other words, if we choose $K$ parameter values \smash{$\thetasc_1,\ldots,\thetasc_K$}, all computed marginal likelihood values must be \emph{equal}, regardless of the individual parameter $\thetasc_k$.
We call this the \emph{self-consistency criterion},
\begin{align}
\dfrac{p(\thetasc)\,p(\Y\given\thetasc)}{p(\thetasc\given \Y)} &= \text{const} \;\forall\thetasc{\in}\,\Theta
    \;\,\Longrightarrow\, \nonumber\\ 
    \dfrac{p(\thetasc_1)\,p(\Y\given\thetasc_1)}{p(\thetasc_1\given \Y)}&=\ldots=\dfrac{p(\thetasc_K)\,p(\Y\given\thetasc_K)}{p(\thetasc_K\given \Y)},
    \label{eq:consistency-thetas}
\end{align}
where $\Theta$ denotes the admissable parameter space and $\thetasc_1,\ldots,\thetasc_K{\in}\,\Theta$ are arbitrary but fixed parameter values.
In this paper, the self-consistency criterion shall be applied to neural posterior and likelihood estimation.
We will first demonstrate that direct constrained optimization is computationally infeasible.
Then, we will propose a straightforward and robust way to integrate the self-consistency criterion into existing neural density estimators.

\subsection{Na\"ive Approach: Direct Constrained Optimization}

We could optimize the NPE~\eqref{eq:loss_npe} or NPLE~\eqref{eq:loss_nple} objective, subject to the self-consistency constraint~\eqref{eq:consistency-thetas}, through Lagrange multipliers. For~\eqref{eq:loss_npe}, this would involve optimizing
\begin{equation}
\begin{aligned}
    L(\phib, \lambda_{1:K}) &= \mathcal{L}_{\text{NPE}}(\phib) + \mathcal{L}(\phib, \lambda_{1:K}),\;\;\;\text{where}\\
    \mathcal{L}(\phib, \lambda_{1:K}) &\coloneqq \sum_{k=1}^{K} \lambda_k \left( \frac{p(\thetasc_k)p(\Y \given \thetasc_k)}{q_\phi(\thetasc_k\given \Y)} - c \right)
\end{aligned}
\end{equation}
with respect to $\phib$ and $\lambda_{1:K}$.
The inconvenience of the unknown constant $c$ could be resolved by %
formulating a relative constraint, that is, choosing
\begin{equation}\label{eq:constraint}
\begin{aligned}
   &\mathcal{L}(\phib, \lambda_{1:K}) \coloneqq \\ &\sum_{k=1}^{K-1} \lambda_k \left( \frac{p(\thetasc_k)p(\Y \given \thetab_k)}{q_\phi(\thetasc_k \given \Y)} - \frac{p(\thetasc_{k+1})p(\Y \given \thetab_{k+1})}{q_\phi(\thetasc_{k+1}\given \Y)} \right).
\end{aligned}
\end{equation}

While this approach is conceptually appealing, such relative constraints are notoriously hard to optimize in practice, which calls for an alternative solution that is more scalable.

\subsection{Variance Penalty and Self-Consistency Loss}
To overcome the problems of the na\"ive approach, we re-frame the constraint as a variance penalty, which simplifies the optimization process and enhances both computational feasibility and robustness compared to direct second order optimization.
First, \citet{koethe2023changeofvariables} observes that a second-order Taylor expansion of Eq.~\ref{eq:loss_npe} yields
\begin{align}\label{eq:taylor}\nonumber
    \text{KL}\big(p(\thetab \given \Y) \, \big\Vert \, q_{\phib}(\thetab\given\Y)\big) &\approx \\ \frac{1}{2p(\Y)^2} &\,\text{Var}_{p(\thetab \given \Y)}\left(\frac{p(\thetab)\,p(\Y \given \thetab)}{q_{\phib}(\thetab \given \Y)}\right).
\end{align}
This approximation implies that the KL divergence between the true and the approximate posterior becomes negligible as the variance of Eq.~\ref{eq:consistency} with respect to the true posterior $p(\thetab \given \Y)$ shrinks to zero.
Moreover, it is clear that minimizing the variance of the marginal likelihood estimator indirectly achieves the effect of the constraint implied by Eq.~\ref{eq:constraint}.
However, directly targeting the variance introduces two new challenges: (i) The true posterior $p(\thetab \given \Y)$ is unknown; and (ii) the argument of $\text{Var}_{p(\thetab \given \Y)}\left[\cdot\right]$ may cause numerical instabilities due to the danger of vanishingly small values in the denominator.

\begin{algorithm}[t]
\caption{Self-consistency loss for finite training.\\
\textcolor[HTML]{414487}{\{I\}: likelihood-based with analytic likelihood}\\
\textcolor[HTML]{22A184}{\{II\}: simulation-based with approximate likelihood}}
\label{alg:self-consistency}
\begin{algorithmic}[1]
\REQUIRE{$N$ training data tuples $\{(\thetab^{(i)}, \Y^{(i)})\}_{i=1}^N$}
\REQUIRE{Number of self-consistency samples $K$}
\FOR{$i=1,\ldots,N$}
\STATE{< compute other losses such as NPE/NPLE loss >}
\FOR{$k=1,\ldots,K$}
    \STATE{$\thetasc_k\sim \pi(\thetab)$} \hfill\COMMENT{sample from proposal $\pi(\thetab)$}
    \STATE{$\log \hat{p}_k(\Y^{(i)}) = 
    \begin{cases}
        \log \dfrac{p(\thetasc_k)\,\textcolor[HTML]{414487}{p(\Y^{(i)}\given\thetasc_k)}}{q_{\phib}(\thetasc_k\given \Y^{(i)})}\;\;\;\text{\textcolor[HTML]{414487}{\{I\}}}\\[16pt]
        \log\dfrac{p(\thetasc_k)\,\textcolor[HTML]{22A184}{q_{\etab}(\Y^{(i)}\given\thetasc_k)}}{q_{\phib}(\thetasc_k\given \Y^{(i)})}\,\text{\textcolor[HTML]{22A184}{\{II\}}}
    \end{cases}
    $}
\ENDFOR
\STATE{$\mathcal{L}_{SC}^{(i)} = \Var(\{\log \hat{p}_k(\Y^{(i)})\}_{k=1}^K)$}
\ENDFOR
\end{algorithmic}
\end{algorithm}

As a remedy, we propose a two-step solution during the simulation-based training: (i) Sample parameters $\thetasc$ from a proposal distribution $\pi(\thetab)$; and (ii) quantify the expected violation of the self-consistency criterion via the variance of the estimated \textit{log} marginal likelihood (LML) across these parameter samples.
This results in the following \emph{self-consistency loss function}:
\begin{equation}\label{eq:sc-loss}
    \mathcal{L}_\text{SC} (\Y,\phib) = \Var_{\pi(\thetab)}\left(\log \frac{ p(\thetasc)\,p(\Y\given \thetasc)}{q_{\phib}(\thetasc\given{\Y})}\right)
\end{equation}
The analytic likelihood $p(\Y\given\thetasc)$ may be replaced with an approximate likelihood $q_{\etab}(\Y\given\thetasc)$, as demonstrated in \textbf{Experiments 2}, \textbf{4}, and \textbf{5}.
If the variance in Eq.~\ref{eq:sc-loss} is zero, the estimated marginal likelihood is constant across the parameter space $\Theta$.
In other words, the approximation is self-consistent (cf.\ \autoref{fig:main-figure}).
The following proposition warrants the functional equivalence between using the variance in Eq.~\ref{eq:taylor} and a tractable version of the more stable formulation in Eq.~\ref{eq:sc-loss}. The proof of the proposition is given in \textbf{Appendix~\ref{app:proofs}}.

\begin{prop}\label{prop:first}
Let $\pi(\thetab)$ be any proposal distribution with the same support as $p(\thetab \given \Y)$, $\Y$ be a fixed data set, and $f$ be any monotonic function, then
\begin{align}\nonumber
    \Var_{\pi(\thetab)}\left(f\left(\frac{p(\Y \given \thetab)\,p(\thetab)}{q(\thetab \given \Y)}\right)\right) &= 0 \implies\\
    \Var_{p(\thetab \given \Y)}\left(\frac{p(\Y \given \thetab)\,p(\thetab)}{q(\thetab \given \Y)}\right)\ &= 0. \nonumber
\end{align}
\end{prop}
This proposition states that (i) minimizing the variance of the log marginal likelihood is equivalent to targeting the correct quantity in Eq.~\ref{eq:taylor}; and (ii) we can take a different proposal than the unknown posterior $p(\thetab \given \Y)$ and still minimize the correct variance term in Eq.~\ref{eq:taylor}.
For example, taking the logarithm of the marginal likelihood is akin to using the Gibbs loss in favor of the marginal likelihood for measuring predictive performance \cite{watanabe2009algebraic}.

However, using a different proposal distribution may significantly change the empirical behavior of the Monte Carlo estimate and exhibit poor pre-asymptotic properties, especially if the proposal $\pi(\thetab)$ has much larger variance than the true posterior $p(\thetab\given\Y)$.
Thus, Section~\ref{sec:sc-estimation} discusses techniques for mitigating such behavior by using optimization schedules.
Finally, the same proposition can be shown to hold when both the posterior and the likelihood in the marginal likelihood computation are replaced with neural surrogates.
As a consequence, we would expect that the (reducible) variance in the doubly approximate marginal likelihood will be larger; still our experiments show measurable benefits even when using an approximate likelihood for estimating self-consistency (see \textbf{Experiments 2, 3}, and \textbf{5}).

\subsection{Monte Carlo Estimation}\label{sec:sc-estimation}

The self-consistency loss $\mathcal{L}_{\text{SC}}$ can be seamlessly added to NPE or NPLE losses.
For instance, using the maximum likelihood loss for NPE with normalizing flows, we obtain:
\begin{equation}\label{eq:consistency-loss-npe}
\begin{aligned}
    \gL_\text{SC-NPE}(\phib) =&~\mathbb{E}_{p(\Y)}\Biggr[
    \underbrace{\mathbb{E}_{p(\thetab \given \Y)}\big[{-}\log q_{\phib}(\thetab\given \Y)\big]}_{\text{NPE loss (on fixed $\Y$)}}\\
    &+ 
    \underbrace{\lambda\Var_{\pi(\thetab)}\left(
        \log \frac{p(\thetasc)\,p(\Y\given \thetasc)}{ q_{\phib}(\thetasc\given \Y)}
    \right)}_{\text{self-consistency loss\;}\mathcal{L}_{\text{SC}}\;\text{with weight\;}\lambda \geq 0}
    \Biggr]
\end{aligned}
\end{equation}
The variance in Eq.~\ref{eq:sc-loss} must be empirically estimated based on finite samples $\{\thetasc_k\}_{k=1}^{K}$ from some proposal distribution $\pi(\thetab)$, as detailed in Algorithm~\ref{alg:self-consistency}.
Due to the probabilistic symmetry of the joint distribution, high-density regions in the approximate posterior have the potential to cause large deviations in the estimated marginal likelihood landscape (cf.\ \autoref{fig:main-figure}).
Consequently, we choose the approximate posterior as the proposal, \smash{$\pi(\theta):=q_{\phib}(\thetab\given \Y)$}, to render the Monte Carlo estimate efficient.

Using the approximate posterior as a proposal has one caveat: \textit{It is spectacularly bad at very early stages of training}.
To mitigate this, we use a schedule $\mu(\cdot)$ that anneals the weight $\lambda$ during training.
The use of progressive annealing has been established as a successful means to control the influence of a potentially unstable approximation loop \citep[e.g., for consistency models,][]{song2023consistency,song2023improved,schmitt2023cmpe}.
Based on our experiments, we recommend choosing the schedule so that the self-consistency loss is inactive at the start of training, $\mu(0)=0$, and its weight $\lambda$ increases as training progresses.

\subsection{Intuition for Benefits of Self-Consistency}

So far, we presented the theoretical foundation of our self-consistency loss, and \autoref{sec:experiments} will empirically demonstrate its effectiveness.
However, one pivotal question remains: \textit{Why} does the self-consistency loss improve inference?

In a nutshell, the self-consistency loss leverages more information for correctly amortizing $p(\thetab \given \Y)$ than isolated NPE or NLE \textit{with the same number of simulated data sets in the training phase} through two strategies.
First, the self-consistency loss informs the learned posterior and likelihood about their relation to each other, and explicitly rewards correct marginal likelihood estimates in addition to the maximum likelihood training objectives of NPE and NLE.
This substantially reduces the space of admissible solutions in the training objective, upon which the correct solution is found via the maximum likelihood loss.
Second, the self-consistency loss uses analytical density information from the Bayesian joint model (i.e., likelihood and prior).
Since the maximum likelihood losses in NPE and NLE already optimize the networks towards correct sampling based on simulator outputs, the self-consistency loss further enhances the effect by penalizing deviations in the approximate posterior density (and approximate likelihood, if learned) in regions not immediately covered by the simulator.

\begin{figure*}[t]
    \centering
    \begin{minipage}{0.69\textwidth}
        \begin{subfigure}[t]{\linewidth}
            \stackanchor{\textbf{A}}{\rotatebox[origin=t]{90}{\;\;\textbf{NPE (baseline)}\;}}
            \includegraphics[width=\linewidth,valign=t]{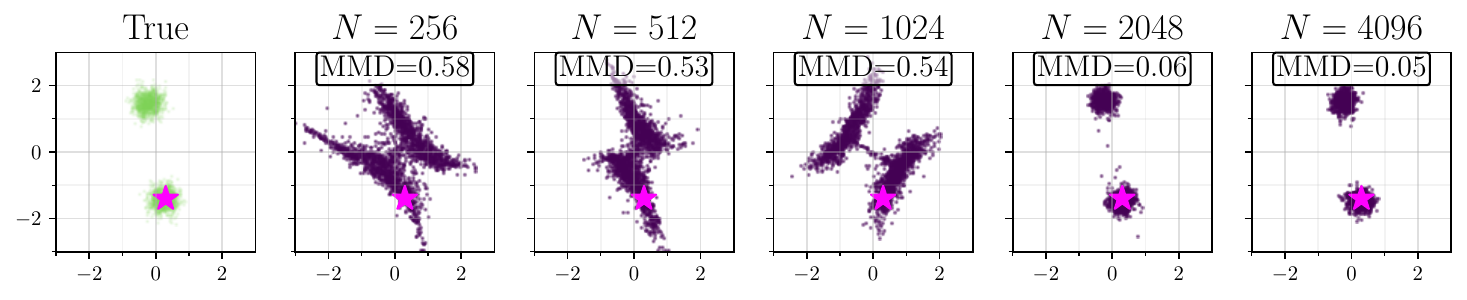}
        \end{subfigure}\\
        \begin{subfigure}[t]{\linewidth}
        \stackanchor{\textbf{B}}{\rotatebox[origin=t]{90}{\textbf{SC-NPE (ours)}\;}}
            \includegraphics[width=\linewidth,valign=t]{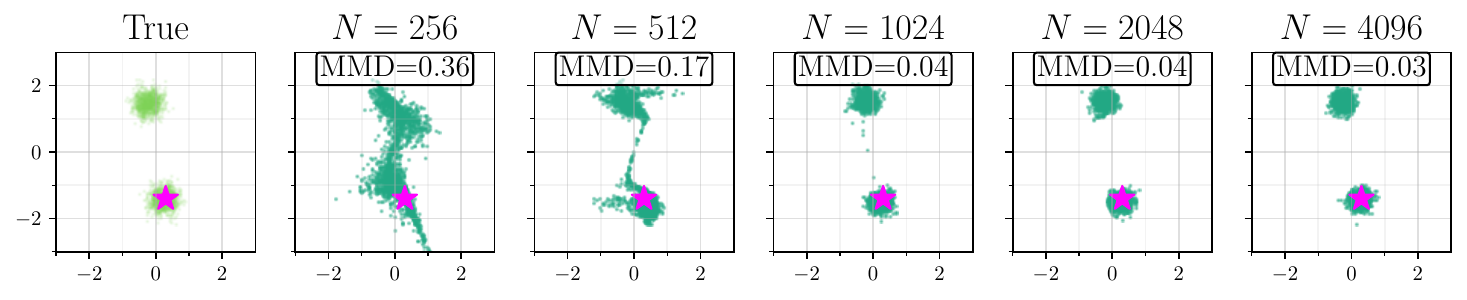}
        \end{subfigure}
    \end{minipage}
    \hfill
    \begin{minipage}{0.28\textwidth}
    \begin{subfigure}[t]{\linewidth}
        \textbf{C}
        \includegraphics[width=1.0\linewidth]{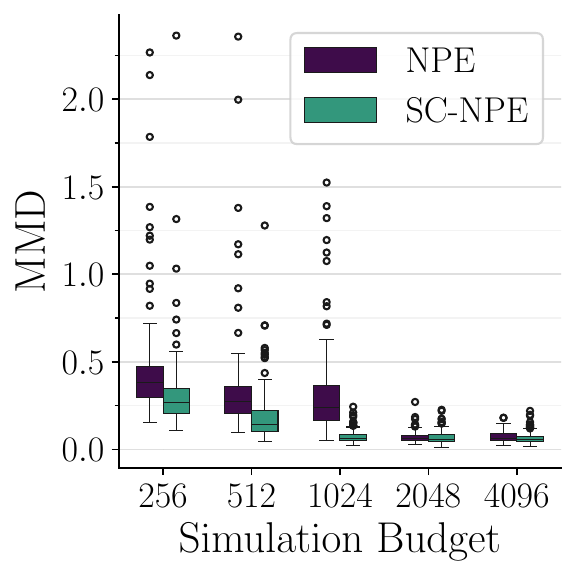}
    \end{subfigure}
    \end{minipage}
    \caption{\textbf{Experiment 1 (Gaussian Mixture Model)}. Performance comparison between the NPE baseline (\textbf{A}) and our self-consistent SC-NPE method (\textbf{B}). %
    Pink star {\color{magenta}$\boldsymbol{\star}$} marks the ground-truth parameter $\thetab^*$.
    Both qualitative assessments (sampling) and quantitative measures (MMD; lower is better) indicate that the our SC-NPE method yields significantly better results given the same neural architecture and training budget.
    Across all simulation budgets, our self-consistent approximator outperforms the NPE baseline, as indexed by improved posterior fidelity (lower MMD) on 100 unseen test instances \textbf{(C)}.
    }
    \label{fig:gmm_main}
\end{figure*}

\section{Related Work}
The self-consistency property in Eq.~\ref{eq:consistency} can be used to compute importance sampling weights during inference to re-weigh the approximate posterior samples \cite{Dax2023importancesampling,glockler2022snvi}.
Importance sampling has been shown to improve the quality of the approximate posteriors when high accuracy is desired.
However, it also increases the necessary amount of computations \emph{during inference}, which in turn impedes low-latency tasks.
In contrast, our self-consistent approximators do not have an increased sampling time during inference.
Similarly, \citet{glockler2022snvi} propose a variational approach to sequential (non-amortized) SBI which entails self-normalized importance sampling based on the marginal likelihood.

\citet{radev2023jointly} use a tandem of two normalizing flows for posterior and likelihood estimation that enables amortized marginal likelihood estimation during \emph{inference}.
Our self-consistency loss additionally applies the principle of amortized marginal likelihood estimation to the neural network \emph{training} phase:
By evaluating the self-consistency of the networks' (log) marginal likelihood estimates, we supply an additional training signal and thereby use the available data more efficiently.
From a different perspective, our self-consistent method with learned likelihoods (SC-NPLE) can be seen as an improvement to jointly amortized neural approximation, which trains the neural approximators in isolation.
In contrast, our self-consistency loss connects the networks during training and explicitly rewards accurate marginal likelihood estimation by the neural network \emph{tandem}.
As such, our work is among only a few approaches that explicitly consider the connection between different quantities in SBI and ABI \citep[for other examples, see][]{brehmer2020mining,chen2023summaries,schmitt2023multinpe}.

A recent review of change-of-variable formulas in generative modeling discusses a multitude of self-consistency properties in (Bayesian) generative models \citep{koethe2023changeofvariables}.
Our self-consistency property of the marginal likelihood in Eq.~\ref{eq:consistency} is one out of multiple possible self-consistency requirements.
The work by \citet{koethe2023changeofvariables} complements the theoretical foundation of our proposed self-consistency loss, which aims to transform their theoretical remarks into a set of actionable methods for amortized Bayesian inference.

\section{Empirical Evaluation}\label{sec:experiments}
We evaluate our self-consistent estimator across a range of synthetic tasks and real-world problems. 
Our main baseline is NPE for tasks with an explicit likelihood, and NPLE for tasks with an implicit likelihood that is learned in tandem.

The key evaluation metric is the maximum mean discrepancy \citep[MMD;][]{Gretton2012} between true and approximate posterior samples.
In addition, we use simulation-based calibration \citep[SBC;][]{talts2018validating,sailynoja2022graphical} to assess the approximators' uncertainty quantification.
Given a true posterior distribution $p(\thetab\given \Y)$, all intervals $U_q(\theta\mid \Y)$ are calibrated for every quantile $q \in (0, 1)$,
\begin{equation}
    q = \iint \mathbf{I}[\theta_*\in U_q(\thetab\given \Y)]\,p(\Y\given\thetab_*)\,p(\thetab_*)\diff\thetab_*\diff \Y,
\end{equation}
with indicator function $\mathbf{I}[\cdot]$ \cite{Burkner2023}.
An approximate posterior may violate this equation, resulting in insufficient calibration.

\subsection{Experiment 1: Gaussian Mixture Model}

We first illustrate our method on a 2-dimensional Gaussian mixture model as described in \citet{geffner2022diffusionsbi}. 
The model consists of two symmetrical, equally weighted components with a shared, known covariance matrix. 
A simulated dataset contains ten independent and identically distributed observations $\Y=\{\y_{j}\}_{j=1}^{10}$, generated by first sampling a parameter $\thetab \sim \mathcal{N}(\thetab \given \mathbf{0}, \mathbf{I})$, and then conditionally sampling each observation as 
\begin{equation}
    \y_j \given \thetab \sim 0.5\,\mathcal{N}(\y \given \theta, \mathbf{I}/2) + 0.5\,\mathcal{N}(\y \given {-}\thetab, \mathbf{I}/2).
\end{equation}

We investigate the effect of simulation budget variations on performance, maintaining a fixed self-consistency sample size of $K=10$ and scaling the simulation budget between $N=256$ and $N=4096$, each time doubling the previous budget.
Both NPE (baseline) and SC-NPE (ours) train an identical neural spline flow architecture \cite{durkan2019neural} for 35 epochs.
We choose a stepwise constant annealing schedule for the self-consistency weight $\lambda$ such that $\lambda=0$ for the first 5 epochs, and $\lambda=1$ for the remaining 30 epochs.
\textbf{Appendix~\ref{app:gmm-details}} contains further training details.

\textbf{Results.} 
Our method demonstrates clear superiority over the baseline, particularly at lower budget levels, as detailed in the following.
\autoref{fig:gmm_main}\textbf{A} and \ref{fig:gmm_main}\textbf{B} give a visual illustration of the posterior distributions for a single dataset. 
\autoref{fig:gmm_main}\textbf{C} shows the distribution of MMDs, computed over 100 test datasets for our self-consistent method against the NPE baseline.
Qualitatively, the samples generated by SC-NPE (ours) are visually closer to the target distribution.
Quantitatively, our self-consistent method achieves substantially lower MMD scores than the NPE baseline, indicating a more accurate approximation.
This underscores the efficiency of integrating self-consistency when learning amortized posteriors, highlighting how our approach improves inference performance with limited computational resources.
All approximators are well-calibrated (see \textbf{Appendix \ref{app:gmm-sbc}}).

\textbf{Ablation: Number of Monte Carlo Samples.}
We vary the number $K\in\{10, 100, 500\}$ of Monte Carlo samples to estimate the variance in the self-consistency loss with NPE \eqref{eq:loss_npe} in \textbf{Appendix~\ref{app:gmm-sc-samples}}.
While we observe satisfactory calibration for all self-consistent architectures, increasing the number of consistency samples beyond $K=10$ does not noticeably improve performance in this experiment.

\textbf{Variation: Approximate Neural Likelihood.} 
We parallel \textbf{Experiment 1} with an approximate likelihood and a simulation budget of $N=1024$ in \textbf{Appendix~\ref{app:gmm-nple}}.
Once again, our self-consistent approximator shows superior performance with respect to density estimation and sampling.

\textbf{Extension: Sequential NPE with Self-Consistency.}
We observe that sequential neural posterior estimation \citep[SNPE;][]{greenberg2019automatic} also benefits from adding our self-consistency loss during training, as evidenced by more accurate posterior samples (see \textbf{Appendix~\ref{app:gmm-sc-snpe}}).
This result further underscores the modularity and flexibility of our proposed self-consistency loss.

\begin{figure}[t]
    \centering
    \includegraphics[width=\linewidth]{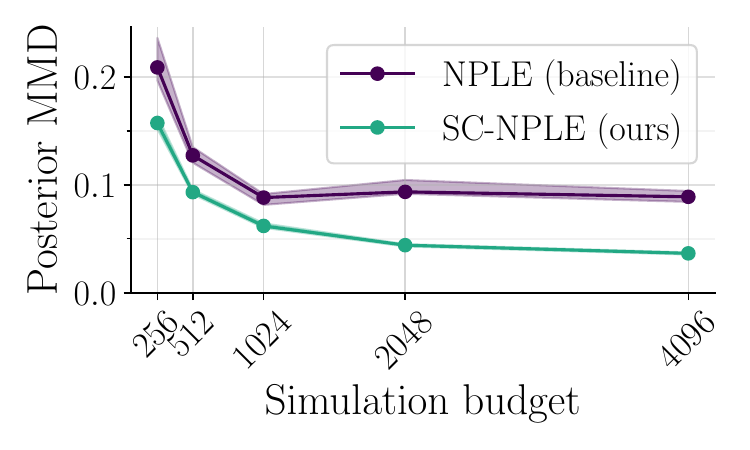}
    \caption{\textbf{Experiment 2 (Two Moons).} Our self-consistency loss yields a lower posterior error (MMD) than the baseline NPLE algorithm on the test set with equal architecture.
    We repeat the experiment 5 times on the same training set; plots show the median, best, and worst run.
    }
    \label{fig:two-moons-simulation-budget}
\end{figure}

\subsection{Experiment 2: Two Moons}
The two moons benchmark is characterized by a bimodal posterior with two crescents, which a posterior approximator needs to recover \cite{greenberg2019automatic, lueckmann2021benchmarking, wiqvist2021sequential, radev2023jointly, schmitt2023cmpe}.
We simultaneously learn an approximate posterior and a surrogate likelihood (i.e., NPLE).
Hence, the self-consistency loss will be completely simulation-based and use the learned surrogate instead of an explicit likelihood (cf.\,Algorithm \ref{alg:self-consistency}, case II).
We repeat the experiment for different training budgets $M\in \{256, 512, 1024, 2048, 4096\}$ to assess the performance under varying data availability.
While $M=256$ is a very small budget for the two moons benchmark, $M=4192$ is generally considered sufficient for this experiment.
We fix the number of Monte Carlo samples to estimate the variance in Eq.~\ref{eq:sc-loss} to $K=10$ and repeat the training loop five times for each architecture to gauge the stochasticity of the training.
We evaluate the approximators' ability to estimate (i) the posterior; (ii) the likelihood; and (iii) the marginal likelihood.
The \textbf{Appendix} contains all neural network training details.

\textbf{Results.} 
Our self-consistent approximator SC-NPLE consistently outperforms the baseline NPLE algorithm with respect to posterior estimation across all simulation budgets, as indexed by a better (lower) average posterior MMD on 100 unseen test instances across 5 training repetitions (see \autoref{fig:two-moons-simulation-budget}).
In this experiment, SC-NPLE (ours) only needs a simulation budget of $M=512$ to perform on-par with the NPLE baseline that was trained on $8\times$ the simulation budget (i.e., $M=4096$).
Further, we observe a more stable training for SC-NPLE: The best and worst training runs in \autoref{fig:two-moons-simulation-budget} have almost equal performance, while the posterior accuracy of NPLE varies between repetitions.
\autoref{tab:tm_likelihood_density} shows the estimated likelihood density of an observed data set $\Y_{\text{real}}$ given the ground-truth parameter $\thetab^*$ which was used to simulate the data set.
While the credible intervals of NPLE (baseline) and SC-NPLE (ours) across the test set have substantial overlap and perform on-par for small simulation budgets, our self-consistent approximator assigns higher (better) likelihood densities to the ground-truth at large simulation budgets.
It is worth noting that the width of the CI is smaller for SC-NPLE, which is a desirable property indicating a reduced approximation error.
Finally, we estimate the marginal likelihood of 500 unseen test instances and observe that the estimates from our self-consistent approximator are significantly sharper (smaller width of the 95\% for fixed data sets), which is an indicator for a reduced approximation error (see \autoref{tab:tm_lml_sharpness}).

\begin{table}[t]
    \centering
    \small
    \setlength\tabcolsep{3.5pt}
    \caption{\textbf{Experiment 2 (Two Moons).} Log likelihood density of the observed data $\Y_{\text{real}}$ under the true data-generating parameter $\thetab^*$ (higher is better).
    We report the mean$\pm$SE across 1000 unseen test instances.}
    \label{tab:tm_likelihood_density}
    \begin{tabular}{l|cccc}
    \toprule
    \textbf{Method} & $N{=}512$ & $N{=}1024$ & $N{=}2048$ & $N{=}4096$\\
    \midrule
    NPLE & $\textbf{3.15} {\pm} 0.03$ & $3.18 {\pm} 0.03$ & $2.88 {\pm} 0.04$ & $2.91 {\pm} 0.05$\\
    SC-NPLE & $3.14 {\pm} 0.02$ & $\textbf{3.45} {\pm} 0.02$ & $\textbf{3.71} {\pm} 0.02$ & $\textbf{3.90} {\pm} 0.02$\\
    \bottomrule
    \end{tabular}
\end{table}

\begin{table}[b]
    \centering
    \small
    \setlength\tabcolsep{3.5pt}
    \caption{\textbf{Experiment 2 (Two Moons).} Approximation error of the log marginal likelihood (LML) estimate (lower is better).
    Our self-consistent estimator yields a significantly smaller approximation error, as indicated by sharper LML estimates.
    For a data set $\Y_{\text{real}}$, the approximation error is quantified as the width of the LML estimate's 95\% CI. We report its mean$\pm$SE across 1000 unseen test instances.
    }
    \label{tab:tm_lml_sharpness}
    \begin{tabular}{lcccc}
    \toprule
    \textbf{Method} & $N{=}512$ & $N{=}1024$ & $N{=}2048$ & $N{=}4096$\\
    \midrule
    NPLE & $6.51 {\pm} 0.11$ & $7.28 {\pm} 0.10$ & $9.07 {\pm} 0.06$ & $10.21 {\pm} 0.08$\\
    SC-NPLE & $\textbf{1.70} {\pm} 0.02$ & $\textbf{1.37} {\pm} 0.02$ & $\textbf{1.21} {\pm} 0.01$ & $\textbf{\;1.14} {\pm} 0.01$\\
    \bottomrule
    \end{tabular}
\end{table}

\begin{figure*}[t]
    \centering
    \begin{subfigure}[t]{0.72\linewidth}
        \stackanchor{\textbf{A}}{\rotatebox[origin=t]{90}{\textbf{NPLE (baseline)}\;}}
        \includegraphics[width=\linewidth,valign=t]{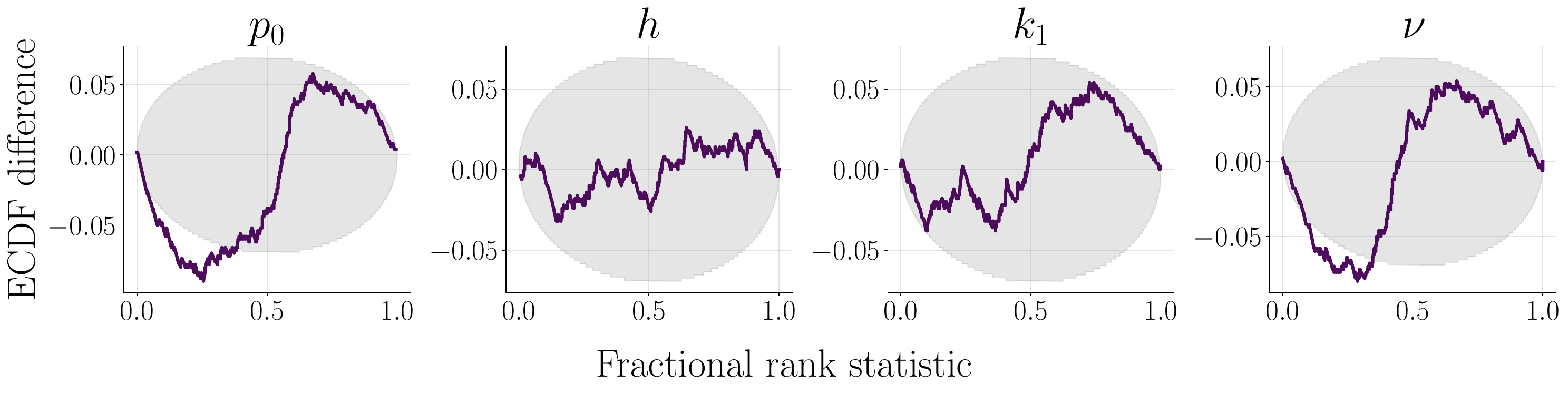}
    \end{subfigure}\hfill
    \begin{subfigure}[t]{0.25\linewidth}
        \textbf{B}
        \includegraphics[width=\linewidth,valign=t]{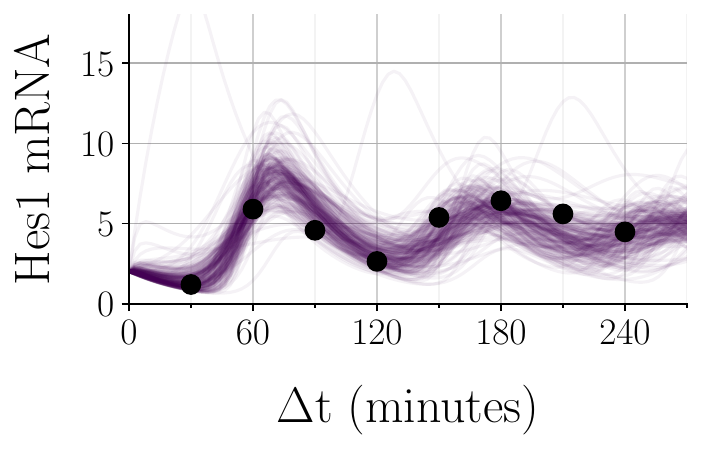}
    \end{subfigure}
    \\
    \begin{subfigure}[t]{0.72\textwidth}
        \stackanchor{\textbf{C}}{\rotatebox[origin=t]{90}{\textbf{SC-NPLE (ours)}\;}}
        \includegraphics[width=\linewidth,valign=t]{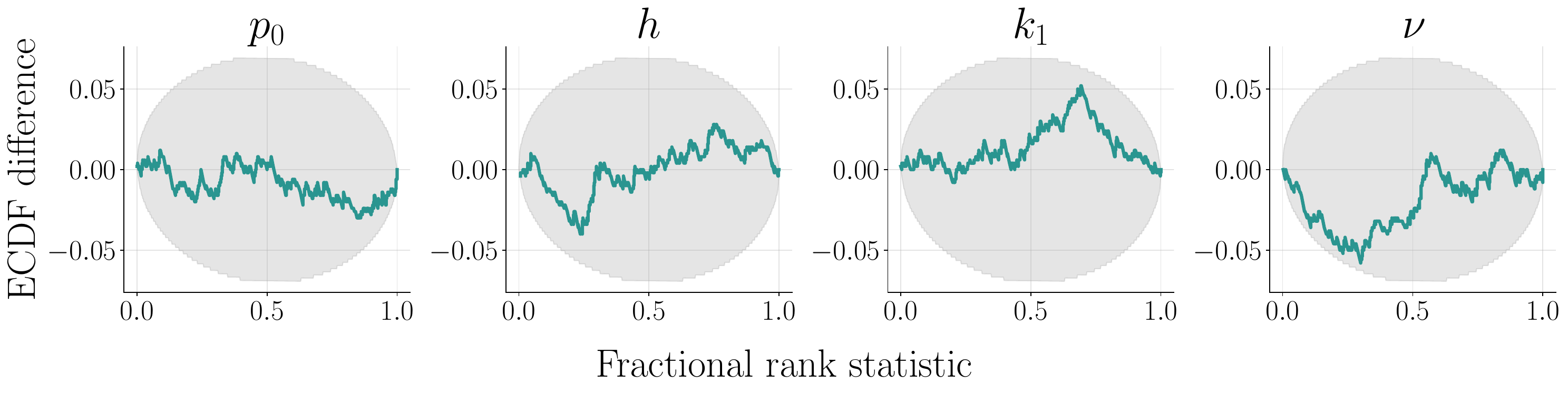}
    \end{subfigure}
    \hfill
    \begin{subfigure}[t]{0.25\linewidth}
    \textbf{D}
    \includegraphics[width=\linewidth,valign=t]{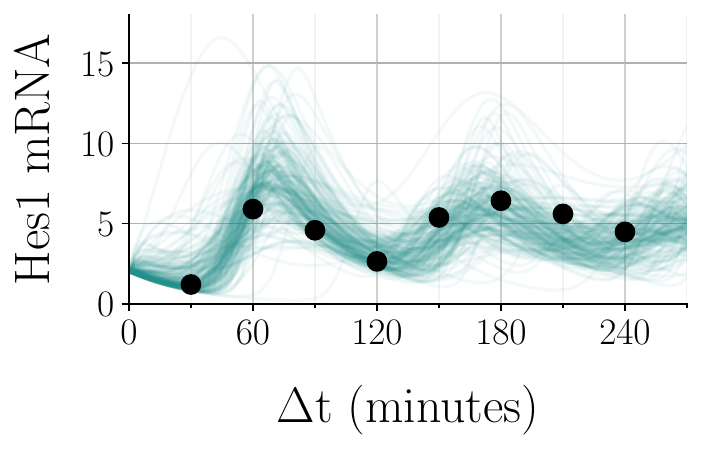}
    \end{subfigure}
    \caption{\textbf{Experiment 3 (Hes1 Expression).} The baseline NPLE approximator shows deficient simulation-based calibration, as indexed by ECDF lines outside the gray 95\% confidence bands \textbf{(A)}. In contrast, our self-consistent approximator is well-calibrated \textbf{(C)}. Samples from the posterior predictive distribution on real experimental data \citep[black dots;][]{Silk2011} are comparable between NPLE \textbf{(B)} and SC \textbf{(D)}.}
    \label{fig:hes1-figure}
\end{figure*}

\subsection{Experiment 3: Oscillatory Hes1 Expression Model}
As a scientific real-world example, we apply our method to an experimental data set in biology \cite{Silk2011}. 
Upon serum stimulation of certain cell lines, the transcription factor Hes1 exhibits sustained oscillatory transcription patterns \cite{Momiji2008}.
The measured concentration of Hes1 mRNA is modeled with a set of three differential equations which are governed by four parameters $\thetab=(p_0, h, k_1, \nu)$ with fixed initial conditions according to \citet{Filippi2011}, see \textbf{Appendix \ref{app:hes1}} for details.

We use a fixed simulation budget of $N=512$ data sets for neural network training.
This simulation budget enables amortized inference, yet it is orders of magnitude smaller than the required budget for approximate Bayesian computation algorithms \citep[e.g., ABC-SMC;][]{Sisson2007} for a single observed data set \cite{Silk2011}.
We train a neural surrogate likelihood in tandem with the neural posterior (NPLE) to use an approximate likelihood $q_{\etab}(\Y\given\thetab)$ in the self-consistency loss.
The self-consistent approximator uses $K=500$ Monte Carlo samples. 
The annealing schedule yields $\lambda=0$ for the first 10 epochs, and $\lambda=1$ for the remaining 60 epochs.
See \textbf{Appendix~\ref{app:hes1}} for details.

\textbf{Results.}
Our self-consistent approximator with approximate likelihood shows superior simulation-based calibration compared to the NPLE baseline, particularly with respect to the parameters $p_0$ and $\nu$ (see \autoref{fig:hes1-figure}\textbf{A} and \textbf{C}).
The posterior predictive distributions of both methods have a similar fit to the real experimental time series $\Y_{\text{real}}$ from \citet{Silk2011}, see \autoref{fig:hes1-figure}\textbf{B} and \ref{fig:hes1-figure}\textbf{D}.

\textbf{Ablation: Underexpressive likelihood network.}
For models where the likelihood is only implicitly defined, an auxiliary likelihood surrogate must be learned. One possible concern is that enforcing self-consistency between the surrogate likelihood and the posterior might actually hurt posterior inference if the likelihood is much more challenging to approximate than the posterior.
We repeat \textbf{Experiment 3} with an underexpressive likelihood network that only features linear units. By ensuring that the likelihood surrogate is insufficient by design, we can emulate situations in which the approximation of the likelihood is unreliable and observe its impact on posterior inference. 
We observe no substantial drop in posterior performance compared to the reference posterior (see \textbf{Appendix}~\ref{app:hes1} for details).
\subsection{Experiment 4: Source Location Finding}

We adapt the source finding experiment from \citet{Foster2021DeepAD}, which
involves finding the location $\thetab$ of a hidden source in 2D. The source emits a signal whose intensity decays inversely with the square of distance. 
We observe a noise corrupted version~$\Y$ of that signal through $N=30$ fixed measurement points.
\begin{table}[b!]
    \caption{\textbf{Experiment 4 (Source Location Finding).} MMD as a function of the simulation budget $N$ and the number $K$ of Monte Carlo samples in Eq.~\ref{eq:sc-loss}. We report mean$\pm$SE across 1000 unseen test instances (lower is better).
    Our self-consistent approximators outperform the NPE baseline throughout all simulation budgets, and the advantage is particularly attenuated at low budgets.
    }
    \label{tab:locfin_budget_scsamples}
    \centering
    \small
    \setlength\tabcolsep{2.3pt}
    \begin{tabular}{llcccc}
    \toprule
            && $N=512$ & $N=1024$ & $N=2048$ & $N=4096$ \\
        \midrule
   \multicolumn{2}{l}{NPE} & 0.54 $ \pm $ 0.02 & 0.39 $ \pm $ 0.02 & 0.25 $ \pm $ 0.02 & 0.22 $ \pm $ 0.01 \\
   
   \multirow{5}{0.2cm}{\rotatebox[origin=t]{90}{SC-NPE (Ours)\,}}& $K{=}5$ & 0.47 $ \pm $ 0.03 & 0.29 $ \pm $ 0.02 & \textbf{0.24} $ \pm $ 0.02 & \textbf{0.20} $ \pm $ 0.02 \\
   & $K{=}10$ & 0.53 $ \pm $ 0.04 & \textbf{0.27} $ \pm $ 0.02 & \textbf{0.24} $ \pm $ 0.02 & 0.22 $ \pm $ 0.02 \\
   & $K{=}20$ & 0.51 $ \pm $ 0.03 & 0.32 $ \pm $ 0.02 & \textbf{0.24} $ \pm $ 0.02 & 0.23 $ \pm $ 0.02 \\
   & $K{=}50$ & 0.52 $ \pm $ 0.04 & 0.30 $ \pm $ 0.02 & 0.26 $ \pm $ 0.02 & 0.21 $ \pm $ 0.02 \\
   & $K{=}100$ & 0.42 $ \pm $ 0.03 & 0.29 $ \pm $ 0.02 & 0.25 $ \pm $ 0.02 & 0.21 $ \pm $ 0.02 \\
   & $K{=}500$ & \textbf{0.40} $ \pm $ 0.03 & 0.32 $ \pm $ 0.02 & 0.25 $ \pm $ 0.02 & 0.21 $ \pm $ 0.02 \\
       \bottomrule
    \end{tabular}
\end{table}
We systematically vary the number of Monte Carlo samples to estimate the variance in Eq.~\ref{eq:sc-loss} as $K\in\{5,10,20,50,100,500\}$. %
Both NPE (baseline) and SC-NPE (ours) use identical neural networks and are trained for 35 epochs on identical settings to ensure a fair comparison.
The annealing schedule for the self-consistency loss weight $\lambda$ is piecewise constant: It yields $\lambda$ for the first 20\% of the training loop, then switching to $\lambda=0.01$.
\textbf{Appendix}~\ref{app:source-location} contains details on the neural network architectures and training scheme.

\textbf{Results.}
Table~\ref{tab:locfin_budget_scsamples} reports the MMD between amortized posterior approximation and a reference posterior from Hamiltonian Monte Carlo \citep[HMC; as implemented in Stan,][]{carpenter2017stan}.
Our self-consistent method demonstrates superior performance to the baseline NPE across all simulation budgets and number of SC samples $K$. The performance advantage is particularly pronounced at lower simulation budgets. Increasing the number of SC samples does not generally result in significantly improved performance.

\subsection{Experiment 5: High-Dimensional Time Series Model}
We demonstrate the effectiveness of the self-consistency loss for high-dimensional data without assuming an explicit likelihood.
To this end, we implement an autoregressive compartmental time series model where the data $\Y$ is a 160-dimensional vector.
We simultaneously learn the posterior and likelihood (NPLE) based on $N=2048$ training examples.
The task of learning the likelihood is directly affected by the high data dimensionality since we learn the likelihood in the uncompressed 160-dimensional data space.
We benchmark standard NPLE against NPLE including our self-consistency loss (SC-NPLE) with $K\in \{5, 10, 50, 100\}$ Monte Carlo samples to estimate the variance in Eq.~\ref{eq:sc-loss}.

\textbf{Results.}
\begin{table}[t!]
    \caption{\textbf{Experiment 5 (High-dimensional time series model).} MMD as a function of the number $K$ of self-consistency Monte Carlo samples and a simulation budget $N=2048$. We report mean$\pm$SE across 50 test instances.
    }
    \vspace{-5pt}
    \label{tab:high_dim_ts}
    \centering
    \small
    \setlength\tabcolsep{2.3pt}
    \begin{tabular}{ccccc}
    \toprule
          \multirow{ 2}{*}{NPLE} & \multicolumn{4}{|c}{SC-NPLE (Ours)}  \\
          & \multicolumn{1}{|c}{$K=5$} & $K=10$ & $K=50$ & $K=100$ \\
        \midrule
    0.88 $ \pm $ 0.06 & 0.78 $\pm$ 0.04 & 0.75 $\pm$ 0.05 & 0.69 $\pm$ 0.04 & \textbf{0.54} $\pm$ 0.03 \\
       \bottomrule
    \end{tabular}
\end{table}
We report the maximum mean discrepancy (MMD) to a HMC reference posterior across 50 unseen test instances. 
Table~\ref{tab:high_dim_ts} shows the results,
demonstrating a clear trend of monotonic performance improvement with an increasing number $K$ of self-consistency Monte Carlo samples.
This demonstrates the effectiveness of our self-consistency loss, even when dealing with high-dimensional data where the likelihood estimation can be notably more challenging.

\section{Conclusion}\label{sec:conclusion}

We proposed a new method to exploit inherent symmetry in a joint probabilistic model $p(\thetab, \Y)$ to improve amortized Bayesian inference.
Across four experiments, we illustrated that the combination of simulation-based inference and (approximate) likelihood-based learning increases the efficiency of neural posterior and likelihood estimation.
Concretely, we demonstrated that an additional self-consistency loss leads to (i) better posterior densities; (ii) better posterior samples; (iii) better likelihood densities; and (iv) sharper marginal likelihood estimates.
The advantage of our self-consistent estimator is particularly evident for low data scenarios, which is a frequent bottleneck in real-world applications of simulation-based inference \citep[e.g.,][]{zhang2023sensitivity, zeng2023probabilistic, bharti2022approximate}

\textbf{Limitations.} As always, there is no free lunch: The improved performance though our self-consistency loss comes with an increased computational cost.
However, the self-consistency loss is designed to pre-pay the cost \emph{during the training stage}.
As a consequence, the inference algorithm remains unaltered and we maintain rapid amortized inference.
Hence, the trade-off is ideal for applied scenarios where the upfront training time is not a bottleneck, but training data is scarce and fast inference is desired.
Further, our method relies on the ability to evaluate the prior density $p(\thetab)$.
This currently limits its applicability to scenarios where the prior density is available in analytic form (most probabilistic modeling applications) or can be learned.

\textbf{Outlook.} While this paper focused on amortized Bayesian inference with normalizing flows, our self-consistency loss can readily be applied to sequential simulation-based inference \cite{papamakarios2019sequential,greenberg2019automatic,glockler2022snvi,wiqvist2021sequential}.
Likewise, other conditional density estimators like score modeling \cite{geffner2022diffusionsbi,sharrock2022diffusionsbi,pacchiardi2022score}, flow-matching \cite{lipman2023flow}, or consistency models \cite{schmitt2023cmpe} may benefit from our additional loss function as well.
Finally, future research could explore variations and extensions of our proposed method, such as different proposal distributions $\pi(\thetab)$ for more efficient Monte Carlo estimates, prior density learning, likelihood learning on summaries instead of the raw data, or improved loss functions altogether that build on the principle of self-consistency.
\textbf{Appendix \ref{app:faq}} contains a selection of Frequently Asked Questions (FAQ) that a reader might have.

\FloatBarrier

\section*{Code Availability}
We provide reproducible code in the open repository at \url{https://github.com/marvinschmitt/self-consistency-abi}

\section*{Impact Statement}
This paper presents work that advances the field of amortized Bayesian inference (ABI) by rendering analyses more data-efficient.
As such, our method can be used to improve the results in malign applications as well, and its societal implications need to be evaluated on an individual basis.

\section*{Acknowledgments}
MS was supported by the Cyber Valley Research Fund (grant number: CyVy-RF-2021-16) and the Deutsche Forschungsgemeinschaft (DFG, German Research Foundation) under Germany’s Excellence Strategy EXC-2075 - 390740016 (the Stuttgart Cluster of Excellence SimTech).
DRI is supported by EPSRC through the Modern Statistics and Statistical Machine Learning (StatML) CDT programme, grant no. EP/S023151/1.
DH was supported by the Deutsche Forschungsgemeinschaft (DFG, German Research Foundation) under project 508399956.
UK was supported by the Informatics for Life initiative funded by the Klaus Tschira Foundation.

\bibliography{references.bib}

\begin{thebibliography}{51}
\providecommand{\natexlab}[1]{#1}
\providecommand{\url}[1]{\texttt{#1}}
\expandafter\ifx\csname urlstyle\endcsname\relax
  \providecommand{\doi}[1]{doi: #1}\else
  \providecommand{\doi}{doi: \begingroup \urlstyle{rm}\Url}\fi

\bibitem[Alexanderson \& Henter(2020)Alexanderson and Henter]{alexanderson2020studentising}
Alexanderson, S. and Henter, G.~E.
\newblock Robust model training and generalisation with studentising flows, 2020.

\bibitem[Avecilla et~al.(2022)Avecilla, Chuong, Li, Sherlock, Gresham, and Ram]{avecilla2022neural}
Avecilla, G., Chuong, J.~N., Li, F., Sherlock, G., Gresham, D., and Ram, Y.
\newblock Neural networks enable efficient and accurate simulation-based inference of evolutionary parameters from adaptation dynamics.
\newblock \emph{PLoS Biology}, 20\penalty0 (5):\penalty0 e3001633, 2022.

\bibitem[Bharti et~al.(2022)Bharti, Filstroff, and Kaski]{bharti2022approximate}
Bharti, A., Filstroff, L., and Kaski, S.
\newblock Approximate {Bayesian} computation with domain expert in the loop.
\newblock In \emph{International Conference on Machine Learning}, pp.\  1893--1905. PMLR, 2022.

\bibitem[Blei et~al.(2017)Blei, Kucukelbir, and McAuliffe]{blei2017variational}
Blei, D.~M., Kucukelbir, A., and McAuliffe, J.~D.
\newblock Variational inference: A review for statisticians.
\newblock \emph{Journal of the American statistical Association}, 112\penalty0 (518):\penalty0 859--877, 2017.

\bibitem[Brehmer et~al.(2018)Brehmer, Cranmer, Louppe, and Pavez]{brehmer2018guide}
Brehmer, J., Cranmer, K., Louppe, G., and Pavez, J.
\newblock A guide to constraining effective field theories with machine learning.
\newblock \emph{Physical Review D}, 98\penalty0 (5):\penalty0 052004, 2018.

\bibitem[Brehmer et~al.(2020{\natexlab{a}})Brehmer, Kling, Espejo, and Cranmer]{brehmer2020madminer}
Brehmer, J., Kling, F., Espejo, I., and Cranmer, K.
\newblock Madminer: Machine learning-based inference for particle physics.
\newblock \emph{Computing and Software for Big Science}, 4:\penalty0 1--25, 2020{\natexlab{a}}.

\bibitem[Brehmer et~al.(2020{\natexlab{b}})Brehmer, Louppe, Pavez, and Cranmer]{brehmer2020mining}
Brehmer, J., Louppe, G., Pavez, J., and Cranmer, K.
\newblock Mining gold from implicit models to improve likelihood-free inference.
\newblock \emph{Proceedings of the National Academy of Sciences}, 117\penalty0 (10):\penalty0 5242--5249, 2020{\natexlab{b}}.

\bibitem[B\"{u}rkner et~al.(2023)B\"{u}rkner, Scholz, and Radev]{Burkner2023}
B\"{u}rkner, P.-C., Scholz, M., and Radev, S.~T.
\newblock Some models are useful, but how do we know which ones? towards a unified bayesian model taxonomy.
\newblock \emph{Statistics Surveys}, 17\penalty0 (none), 2023.
\newblock ISSN 1935-7516.
\newblock \doi{10.1214/23-ss145}.

\bibitem[Carpenter et~al.(2017)Carpenter, Gelman, Hoffman, Lee, Goodrich, Betancourt, Brubaker, Guo, Li, and Riddell]{carpenter2017stan}
Carpenter, B., Gelman, A., Hoffman, M.~D., Lee, D., Goodrich, B., Betancourt, M., Brubaker, M., Guo, J., Li, P., and Riddell, A.
\newblock Stan: A probabilistic programming language.
\newblock \emph{Journal of statistical software}, 76\penalty0 (1), 2017.

\bibitem[Chen et~al.(2023)Chen, Gutmann, and Weller]{chen2023summaries}
Chen, Y., Gutmann, M.~U., and Weller, A.
\newblock Is learning summary statistics necessary for likelihood-free inference?
\newblock In Krause, A., Brunskill, E., Cho, K., Engelhardt, B., Sabato, S., and Scarlett, J. (eds.), \emph{Proceedings of the 40th International Conference on Machine Learning}, volume 202 of \emph{Proceedings of Machine Learning Research}, pp.\  4529--4544. PMLR, 2023.

\bibitem[Cranmer et~al.(2020)Cranmer, Brehmer, and Louppe]{cranmer2020frontier}
Cranmer, K., Brehmer, J., and Louppe, G.
\newblock The frontier of simulation-based inference.
\newblock \emph{Proceedings of the National Academy of Sciences}, 2020.

\bibitem[Dax et~al.(2023)Dax, Green, Gair, P\"{u}rrer, Wildberger, Macke, Buonanno, and Sch\"{o}lkopf]{Dax2023importancesampling}
Dax, M., Green, S.~R., Gair, J., P\"{u}rrer, M., Wildberger, J., Macke, J.~H., Buonanno, A., and Sch\"{o}lkopf, B.
\newblock Neural importance sampling for rapid and reliable gravitational-wave inference.
\newblock \emph{Physical Review Letters}, 130\penalty0 (17), April 2023.
\newblock ISSN 1079-7114.
\newblock \doi{10.1103/physrevlett.130.171403}.

\bibitem[Durkan et~al.(2019)Durkan, Bekasov, Murray, and Papamakarios]{durkan2019neural}
Durkan, C., Bekasov, A., Murray, I., and Papamakarios, G.
\newblock Neural spline flows.
\newblock \emph{Advances in neural information processing systems}, 32, 2019.

\bibitem[Filippi et~al.(2011)Filippi, Barnes, Cornebise, and Stumpf]{Filippi2011}
Filippi, S., Barnes, C., Cornebise, J., and Stumpf, M. P.~H.
\newblock On optimality of kernels for approximate bayesian computation using sequential monte carlo, June 2011.

\bibitem[Foster et~al.(2021)Foster, Ivanova, Malik, and Rainforth]{Foster2021DeepAD}
Foster, A., Ivanova, D.~R., Malik, I., and Rainforth, T.
\newblock Deep adaptive design: Amortizing sequential bayesian experimental design.
\newblock In \emph{International Conference on Machine Learning}, 2021.

\bibitem[Geffner et~al.(2023)Geffner, Papamakarios, and Mnih]{geffner2022diffusionsbi}
Geffner, T., Papamakarios, G., and Mnih, A.
\newblock Compositional score modeling for simulation-based inference.
\newblock In Krause, A., Brunskill, E., Cho, K., Engelhardt, B., Sabato, S., and Scarlett, J. (eds.), \emph{Proceedings of the 40th International Conference on Machine Learning}, volume 202 of \emph{Proceedings of Machine Learning Research}, pp.\  11098--11116. PMLR, 23--29 Jul 2023.

\bibitem[Gelman et~al.(2013)Gelman, Carlin, Stern, Dunson, Vehtari, and Rubin]{BDA3}
Gelman, A., Carlin, J.~B., Stern, H.~S., Dunson, D.~B., Vehtari, A., and Rubin, D.~B.
\newblock \emph{{B}ayesian Data Analysis (3rd Edition)}.
\newblock Chapman and Hall/CRC, 2013.

\bibitem[Gl{\"o}ckler et~al.(2022)Gl{\"o}ckler, Deistler, and Macke]{glockler2022snvi}
Gl{\"o}ckler, M., Deistler, M., and Macke, J.~H.
\newblock Variational methods for simulation-based inference.
\newblock In \emph{International Conference on Learning Representations}, 2022.

\bibitem[Gon{\c{c}}alves et~al.(2020)Gon{\c{c}}alves, Lueckmann, Deistler, et~al.]{gonccalves2020training}
Gon{\c{c}}alves, P.~J., Lueckmann, J.-M., Deistler, M., et~al.
\newblock Training deep neural density estimators to identify mechanistic models of neural dynamics.
\newblock \emph{Elife}, 2020.

\bibitem[Greenberg et~al.(2019)Greenberg, Nonnenmacher, and Macke]{greenberg2019automatic}
Greenberg, D., Nonnenmacher, M., and Macke, J.
\newblock Automatic posterior transformation for likelihood-free inference.
\newblock In \emph{International Conference on Machine Learning}, 2019.

\bibitem[Gretton et~al.(2012)Gretton, Borgwardt, Rasch, Schölkopf, and Smola]{Gretton2012}
Gretton, A., Borgwardt, K., Rasch, M., Schölkopf, B., and Smola, A.
\newblock {A Kernel Two-Sample Test}.
\newblock \emph{The Journal of Machine Learning Research}, 13:\penalty0 723--773, 2012.

\bibitem[Hüllermeier \& Waegeman(2021)Hüllermeier and Waegeman]{hullermeier_aleatoric_2021}
Hüllermeier, E. and Waegeman, W.
\newblock Aleatoric and {Epistemic} {Uncertainty} in {Machine} {Learning}: {An} {Introduction} to {Concepts} and {Methods}.
\newblock \emph{Machine Learning}, 110\penalty0 (3):\penalty0 457--506, March 2021.
\newblock ISSN 0885-6125, 1573-0565.
\newblock \doi{10.1007/s10994-021-05946-3}.

\bibitem[Köthe(2023)]{koethe2023changeofvariables}
Köthe, U.
\newblock A review of change of variable formulas for generative modeling, 2023.

\bibitem[Lavin et~al.(2021)Lavin, Zenil, Paige, et~al.]{lavin2021simulation}
Lavin, A., Zenil, H., Paige, B., et~al.
\newblock Simulation intelligence: Towards a new generation of scientific methods.
\newblock \emph{arXiv preprint}, 2021.

\bibitem[Lee et~al.(2019)Lee, Lee, Kim, Kosiorek, Choi, and Teh]{lee2019settransformer}
Lee, J., Lee, Y., Kim, J., Kosiorek, A., Choi, S., and Teh, Y.~W.
\newblock Set transformer: A framework for attention-based permutation-invariant neural networks.
\newblock In Chaudhuri, K. and Salakhutdinov, R. (eds.), \emph{Proceedings of the 36th International Conference on Machine Learning}, volume~97 of \emph{Proceedings of Machine Learning Research}, pp.\  3744--3753. PMLR, 2019.

\bibitem[Lipman et~al.(2023)Lipman, Chen, Ben-Hamu, Nickel, and Le]{lipman2023flow}
Lipman, Y., Chen, R. T.~Q., Ben-Hamu, H., Nickel, M., and Le, M.
\newblock Flow matching for generative modeling.
\newblock In \emph{The 11th International Conference on Learning Representations}, 2023.

\bibitem[Lueckmann et~al.(2021)Lueckmann, Boelts, Greenberg, Goncalves, and Macke]{lueckmann2021benchmarking}
Lueckmann, J.-M., Boelts, J., Greenberg, D., Goncalves, P., and Macke, J.
\newblock Benchmarking simulation-based inference.
\newblock In Banerjee, A. and Fukumizu, K. (eds.), \emph{Proceedings of The 24th International Conference on Artificial Intelligence and Statistics}, volume 130 of \emph{Proceedings of Machine Learning Research}, pp.\  343--351. PMLR, 13--15 Apr 2021.

\bibitem[Meng \& Wong(1996)Meng and Wong]{meng_simulating_1996}
Meng, X.-L. and Wong, W.~H.
\newblock Simulating ratios of normalizing constants via a simple identity: a theoretical exploration.
\newblock \emph{Statistica Sinica}, 1996.

\bibitem[Miyato et~al.(2018)Miyato, Kataoka, Koyama, and Yoshida]{miyato2018spectral}
Miyato, T., Kataoka, T., Koyama, M., and Yoshida, Y.
\newblock Spectral normalization for generative adversarial networks.
\newblock In \emph{International Conference on Learning Representations}, 2018.

\bibitem[Momiji \& Monk(2008)Momiji and Monk]{Momiji2008}
Momiji, H. and Monk, N.~A.
\newblock Dissecting the dynamics of the hes1 genetic oscillator.
\newblock \emph{Journal of Theoretical Biology}, 254\penalty0 (4):\penalty0 784--798, oct 2008.
\newblock \doi{10.1016/j.jtbi.2008.07.013}.

\bibitem[Neal(2011)]{neal_mcmc_2011}
Neal, R.~M.
\newblock \emph{{MCMC} using {Hamiltonian} dynamics}.
\newblock May 2011.
\newblock \doi{10.1201/b10905}.

\bibitem[Pacchiardi \& Dutta(2022)Pacchiardi and Dutta]{pacchiardi2022score}
Pacchiardi, L. and Dutta, R.
\newblock Score matched neural exponential families for likelihood-free inference.
\newblock \emph{J. Mach. Learn. Res.}, 23:\penalty0 38--1, 2022.

\bibitem[Papamakarios et~al.(2019)Papamakarios, Sterratt, and Murray]{papamakarios2019sequential}
Papamakarios, G., Sterratt, D., and Murray, I.
\newblock Sequential neural likelihood: Fast likelihood-free inference with autoregressive flows.
\newblock In \emph{The 22nd International Conference on Artificial Intelligence and Statistics}, pp.\  837--848. PMLR, 2019.

\bibitem[Radev et~al.(2020)Radev, Mertens, Voss, Ardizzone, and K{\"o}the]{radev2020bayesflow}
Radev, S.~T., Mertens, U.~K., Voss, A., Ardizzone, L., and K{\"o}the, U.
\newblock {BayesFlow}: Learning complex stochastic models with invertible neural networks.
\newblock \emph{IEEE transactions on neural networks and learning systems}, 2020.

\bibitem[Radev et~al.(2023)Radev, Schmitt, Pratz, Picchini, K\"othe, and B\"urkner]{radev2023jointly}
Radev, S.~T., Schmitt, M., Pratz, V., Picchini, U., K\"othe, U., and B\"urkner, P.-C.
\newblock {JANA: Jointly Amortized Neural Approximation of Complex Bayesian Models}.
\newblock In Evans, R.~J. and Shpitser, I. (eds.), \emph{Proceedings of the 39th Conference on Uncertainty in Artificial Intelligence}, volume 216 of \emph{Proceedings of Machine Learning Research}, pp.\  1695--1706. PMLR, 2023.

\bibitem[Rezende \& Mohamed(2015)Rezende and Mohamed]{rezende2015}
Rezende, D.~J. and Mohamed, S.
\newblock Variational inference with normalizing flows.
\newblock In \emph{Proceedings of the 32nd International Conference on International Conference on Machine Learning - Volume 37}, ICML'15, pp.\  1530–1538. JMLR.org, 2015.

\bibitem[S{\"a}ilynoja et~al.(2022)S{\"a}ilynoja, B{\"u}rkner, and Vehtari]{sailynoja2022graphical}
S{\"a}ilynoja, T., B{\"u}rkner, P.-C., and Vehtari, A.
\newblock Graphical test for discrete uniformity and its applications in goodness-of-fit evaluation and multiple sample comparison.
\newblock \emph{Statistics and Computing}, 32\penalty0 (2):\penalty0 1--21, 2022.

\bibitem[Schmitt et~al.(2023{\natexlab{a}})Schmitt, Pratz, Köthe, Bürkner, and Radev]{schmitt2023cmpe}
Schmitt, M., Pratz, V., Köthe, U., Bürkner, P.-C., and Radev, S.~T.
\newblock Consistency models for scalable and fast simulation-based inference, 2023{\natexlab{a}}.

\bibitem[Schmitt et~al.(2023{\natexlab{b}})Schmitt, Radev, and Bürkner]{schmitt2023multinpe}
Schmitt, M., Radev, S.~T., and Bürkner, P.-C.
\newblock Fuse it or lose it: Deep fusion for multimodal simulation-based inference, 2023{\natexlab{b}}.

\bibitem[Sharrock et~al.(2022)Sharrock, Simons, Liu, and Beaumont]{sharrock2022diffusionsbi}
Sharrock, L., Simons, J., Liu, S., and Beaumont, M.
\newblock Sequential neural score estimation: Likelihood-free inference with conditional score based diffusion models, 2022.

\bibitem[Silk et~al.(2011)Silk, Kirk, Barnes, Toni, Rose, Moon, Dallman, and Stumpf]{Silk2011}
Silk, D., Kirk, P.~D., Barnes, C.~P., Toni, T., Rose, A., Moon, S., Dallman, M.~J., and Stumpf, M.~P.
\newblock Designing attractive models via automated identification of chaotic and oscillatory dynamical regimes.
\newblock \emph{Nature Communications}, 2\penalty0 (1), 2011.
\newblock \doi{10.1038/ncomms1496}.

\bibitem[Sisson et~al.(2007)Sisson, Fan, and Tanaka]{Sisson2007}
Sisson, S.~A., Fan, Y., and Tanaka, M.~M.
\newblock Sequential monte carlo without likelihoods.
\newblock \emph{Proceedings of the National Academy of Sciences}, 104\penalty0 (6):\penalty0 1760--1765, February 2007.
\newblock \doi{10.1073/pnas.0607208104}.

\bibitem[Song \& Dhariwal(2023)Song and Dhariwal]{song2023improved}
Song, Y. and Dhariwal, P.
\newblock Improved {{Techniques}} for {{Training Consistency Models}}, October 2023.

\bibitem[Song et~al.(2023)Song, Dhariwal, Chen, and Sutskever]{song2023consistency}
Song, Y., Dhariwal, P., Chen, M., and Sutskever, I.
\newblock Consistency models.
\newblock In \emph{International Conference on Machine Learning}, 2023.

\bibitem[Talts et~al.(2018)Talts, Betancourt, Simpson, Vehtari, and Gelman]{talts2018validating}
Talts, S., Betancourt, M., Simpson, D., Vehtari, A., and Gelman, A.
\newblock Validating {Bayesian} inference algorithms with simulation-based calibration.
\newblock \emph{arXiv preprint}, 2018.

\bibitem[Vehtari et~al.(2022)Vehtari, Gabry, Magnusson, Yao, Bürkner, Paananen, and Gelman]{vehtari2022loo}
Vehtari, A., Gabry, J., Magnusson, M., Yao, Y., Bürkner, P.-C., Paananen, T., and Gelman, A.
\newblock loo: Efficient leave-one-out cross-validation and {WAIC} for {Bayesian} models, 2022.

\bibitem[Watanabe(2009)]{watanabe2009algebraic}
Watanabe, S.
\newblock \emph{Algebraic geometry and statistical learning theory}, volume~25.
\newblock Cambridge university press, 2009.

\bibitem[Wiqvist et~al.(2021)Wiqvist, Frellsen, and Picchini]{wiqvist2021sequential}
Wiqvist, S., Frellsen, J., and Picchini, U.
\newblock Sequential neural posterior and likelihood approximation.
\newblock \emph{arXiv preprint}, 2021.

\bibitem[Zaheer et~al.(2017)Zaheer, Kottur, Ravanbakhsh, Poczos, Salakhutdinov, and Smola]{zaheer2017deepsets}
Zaheer, M., Kottur, S., Ravanbakhsh, S., Poczos, B., Salakhutdinov, R., and Smola, A.
\newblock Deep sets, 2017.

\bibitem[Zeng et~al.(2023)Zeng, Todd, and Hu]{zeng2023probabilistic}
Zeng, J., Todd, M.~D., and Hu, Z.
\newblock Probabilistic damage detection using a new likelihood-free {Bayesian} inference method.
\newblock \emph{Journal of Civil Structural Health Monitoring}, 13\penalty0 (2-3):\penalty0 319--341, 2023.

\bibitem[Zhang \& Mikelsons(2023)Zhang and Mikelsons]{zhang2023sensitivity}
Zhang, Y. and Mikelsons, L.
\newblock Sensitivity-guided iterative parameter identification and data generation with {BayesFlow} and {PELS-VAE} for model calibration.
\newblock \emph{Advanced Modeling and Simulation in Engineering Sciences}, 10\penalty0 (1):\penalty0 1--28, 2023.

\end{thebibliography}
\bibliographystyle{icml2024}

\clearpage
\appendix
\onecolumn
\begin{center}
    {\Large\bfseries\scshape APPENDIX}
\end{center}

\section{Frequently Asked Questions (FAQ)}\label{app:faq}

\textbf{Q: How can I reproduce the results?}\\[3pt]
The code is hosted in a repository at \url{https://github.com/marvinschmitt/self-consistency-abi}

\textbf{Q: How does the self-consistency loss change my current amortized Bayesian workflow?}\\[3pt]
If you have access to an explicit likelihood, you need to implement a \texttt{likelihood.log\_prob} method to evaluate $p(\Y\given\thetasc)$.
This is straightforward with common frameworks such as \texttt{tensorflow\_probability} or \texttt{scipy.stats}.
If your likelihood is implicit (fully simulation-based), your neural likelihood approximator needs to yield a tractable density (e.g., through a normalizing flow).
Further, you need to implement a \texttt{prior.log\_prob} method for your prior distribution, regardless of whether you use NPE (explicit likelihood) or NPLE (implicit likelihood). 
Alternatively, you may try to learn the prior density with an unconditional density estimator on-the-fly (see \autoref{sec:conclusion}).

\textbf{Q: When is it useful to add the self-consistency loss?}\\[3pt]
When the simulation program is computationally costly or the simulation budget is fixed, our experiments suggest that adding the self-consistency loss to an optimization objective might help get more out of the available training data.

\textbf{Q: What about learned summary statistics?}\\[3pt]
Our method is fully compatible with end-to-end learning of summary statistics alongside the neural approximator \cite{radev2020bayesflow,radev2023jointly}.
In fact, \textbf{Experiment 1} uses a DeepSet \cite{zaheer2017deepsets} and \textbf{Experiment 4} uses a SetTransformer \cite{lee2019settransformer} to learn fixed-length summary statistics $h(\Y)$ from the observables $\Y$, which are then passed to the posterior approximator.
As mentioned in \autoref{sec:conclusion}, the likelihood 

\textbf{Q: When do I activate the self-consistency loss during training?}\\[3pt]
This depends on the complexity of the problem. Your approximate posterior (and approximate likelihood, if applicable) should be sufficiently good so that (i) the proposals for \smash{$\tilde{\theta}$} cover relevant regions; and (ii) the log density estimates for the posterior (and likelihood, if applicable) have an acceptable quality for the Monte Carlo estimate in Eq.~\ref{eq:sc-loss}.
Further, you have freedom in designing an annealing schedule $\lambda=\mu(\cdot)$ for the weight of the self-consistency term in the loss function.
For instance, you might opt for a smooth schedule which gradually increases the self-consistency weight.

\textbf{Q: Why aren't the posterior samples in \textbf{Experiment 1} perfectly aligned with the true parameter?}\\[3pt]
The simulated data sets in the Gaussian mixture model only consist of ten observations from the Gaussian mixture model with locations $\thetab$ and ${-}\thetab$.
Due to aleatoric uncertainty in the data-generating process \cite{hullermeier_aleatoric_2021}, the empirical information in the sample does not even suffice to inform the true posterior to concentrate on the true data-generating parameter $\thetab$.
Instead, the goal of an approximate posterior is to match the true posterior including the uncertainty it encodes.

\textbf{Q: Why do the NPE posterior samples look so bad in Experiment 1 at $N=1024$? Other papers show better sampling for NPE at this budget.}\\[3pt]
We observed that NPE shows very inconsistent performance across the parameter space at low simulation budgets.
In other words, NPE performs reasonably well for some parameter regions, and atrociously bad for others.
In \autoref{fig:gmm_main}\textbf{C}, this phenomenon manifests as a large range in the corresponding boxplot, which shows the MMD across different test instances.
In contrast, our SC-NPE method achieves remarkably good performance across the entire parameter space at a simulation budget of $N=1024$.

\textbf{Q: What about non-amortized sequential algorithms, like Sequential Neural Posterior Estimation (SNPE)?}\\[3pt]
Our method can readily be integrated with sequential SBI algorithms, such as SNPE \citep{greenberg2019automatic}.
We observed performance increases when SNPE is equipped with our self-consistency loss (see \textbf{Appendix~\ref{app:gmm-sc-snpe}}).

\clearpage

\section{Proof of Proposition 1}
\label{app:proofs}

In the following, we provide a proof of Proposition~\ref{prop:first}. 

\begin{proof}
First, we are going to make use of the well-known fact that the variance of a constant is zero, and zero variance implies a constant argument, that is, for any continuous random vector $X$, we have
\begin{align}
    f(X) = c  &\implies \Var(f(X)) = 0 \label{eq:first}\\
    \Var(f(X)) = 0 &\implies f(X) = c \label{eq:second}
\end{align}

For (\ref{eq:first}), we simply apply the definition of the variance
\begin{align}
    \Var(c) &= \mathbb{E}\left[ \left(c - \mathbb{E}\left[c\right] \right) ^2\right] \\ 
    &= \mathbb{E}\left[(c - c)^2\right] = 0,
\end{align}
where we used the fact that the expectation of a constant recovers the constant itself, that is, $\mathbb{E}\left[c\right] = c$.

For (\ref{eq:second}), we first note that due to the definition of the variance,
\begin{equation}
    \Var(f(X)) = \mathbb{E}\left[ \left( f(X) - \mathbb{E}\left[f(X)\right] \right)^2\right],
\end{equation}
the squared difference $\left( f(X) - \mathbb{E}\left[f(X)\right] \right)^2$ is strictly positive, so the only way for the variance to become zero is if  $\left( f(X) - \mathbb{E}\left[f(X)\right] \right)^2 = 0$. This, in turn, implies that $f(X) = \mathbb{E}\left[f(X)\right]$ for any realization of $X$, which can only happen if $f(X) = c$.

The proof of Proposition~\ref{prop:first} is structured in two parts. 
First, we show that zero variance of a monotone function directly implies zero variance for the argument.
Second, we show that zero variance with respect to a distribution implies zero variance with respect to a different distribution with the same support.

For the first part, given the assumption of zero variance,
\begin{equation}
   \Var_{p(\thetab \given \Y)}\left(f\left(\frac{p(\Y \given \thetab)p(\thetab)}{q(\thetab \given \Y)}\right)\right) = 0,
\end{equation}
it follows that $f(p(\Y \given \thetab)p(\thetab) / q(\thetab \given \Y)) = c,\,\forall \thetab \in \Theta$.
Since we assume that $f$ is monotone, then the functional argument of $f(\cdot)$ must be constant over $\Theta$, and so
\begin{equation}
   \Var_{p(\thetab \given \Y)}\left(\frac{p(\Y \given \thetab)p(\thetab)}{q(\thetab \given \Y)}\right) = 0.
\end{equation}
For the second part, we again observe that the assumption
\begin{equation}
   \Var_{\pi(\thetab)}\left(f\left(\frac{p(\Y \given \thetab)p(\thetab)}{q(\thetab \given \Y)}\right)\right) = 0
\end{equation}
implies that $f(p(\Y \given \thetab)p(\thetab) / q(\thetab \given \Y)) = c,\,\forall \thetab \in \Theta$. 
Since, by assumption, $\pi(\thetab)$ and $p(\thetab \given \Y)$ have the same support, it follows that
\begin{equation}
   \Var_{p(\thetab \given \Y)}\left(f\left(\frac{p(\Y \given \thetab)p(\thetab)}{q(\thetab \given \Y)}\right)\right) = 0
\end{equation}
Combining this with the previous result from the first part concludes the proof.
\end{proof}
\section{Details about Experiment 1}\label{app:gmm-details}
The normalizing flow uses a heavy-tailed Student-$t$ latent distribution with 100 degrees of freedom to provide a more robust latent space \cite{alexanderson2020studentising}.
The neural spline flow has 4 coupling layers and learnable permutation layers.
As a summary network, a DeepSet \cite{zaheer2017deepsets} learns 4-dimensional embeddings that lift the $i.i.d.$ structure of the exchangeable data $\Y$ and are maximally informative for posterior inference \cite{radev2020bayesflow}.
The neural networks are trained for a total of 35 epochs with a batch size of 32 and an initial learning rate of $10^{-3}$.

\subsection{Calibration}\label{app:gmm-sbc} 
\begin{figure}[H]
    \centering
    \begin{tabular}{ccc}
         NPE & &SC10 \\
         \includegraphics[width=0.40\linewidth]{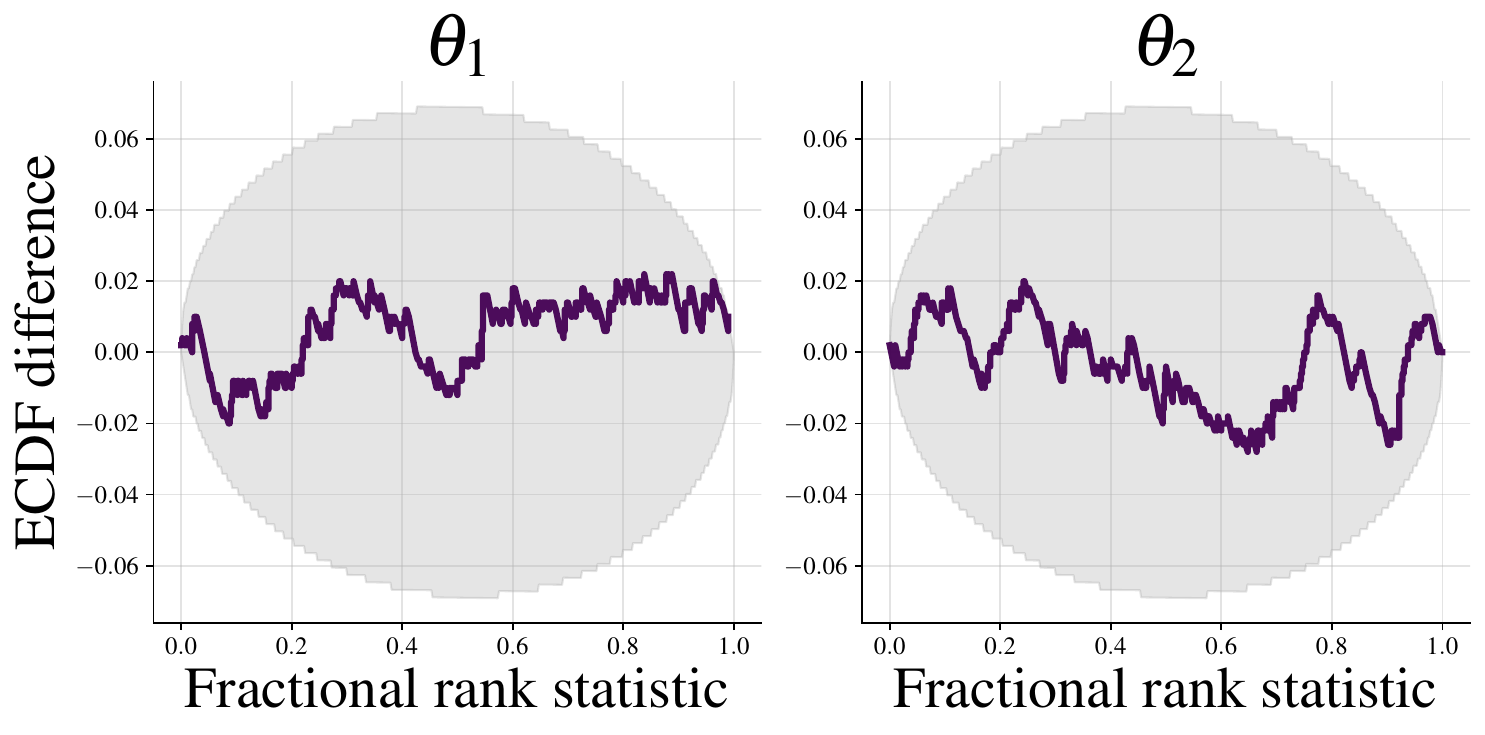}&
         &
         \includegraphics[width=0.40\linewidth]{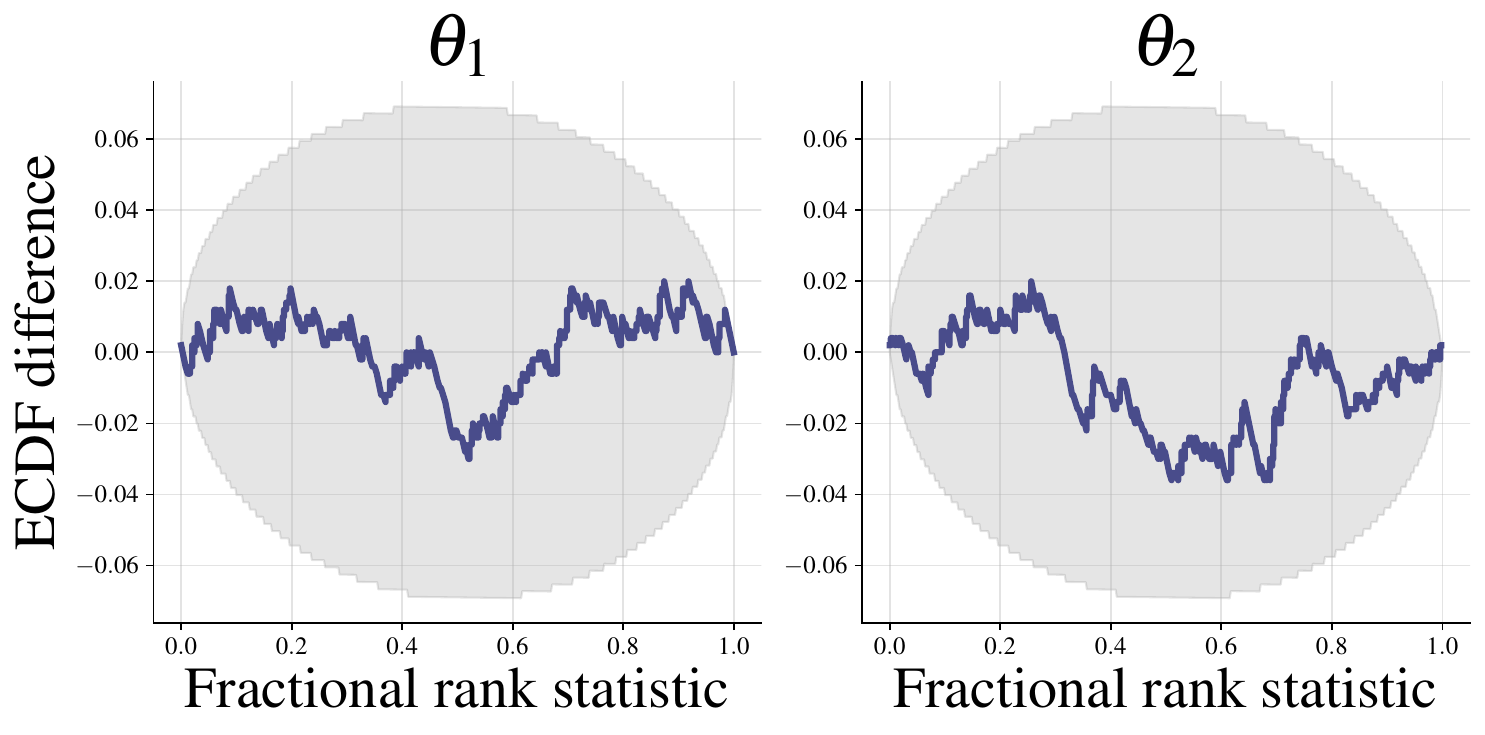} \\
         SC100 & & SC500\\
         \includegraphics[width=0.40\linewidth]{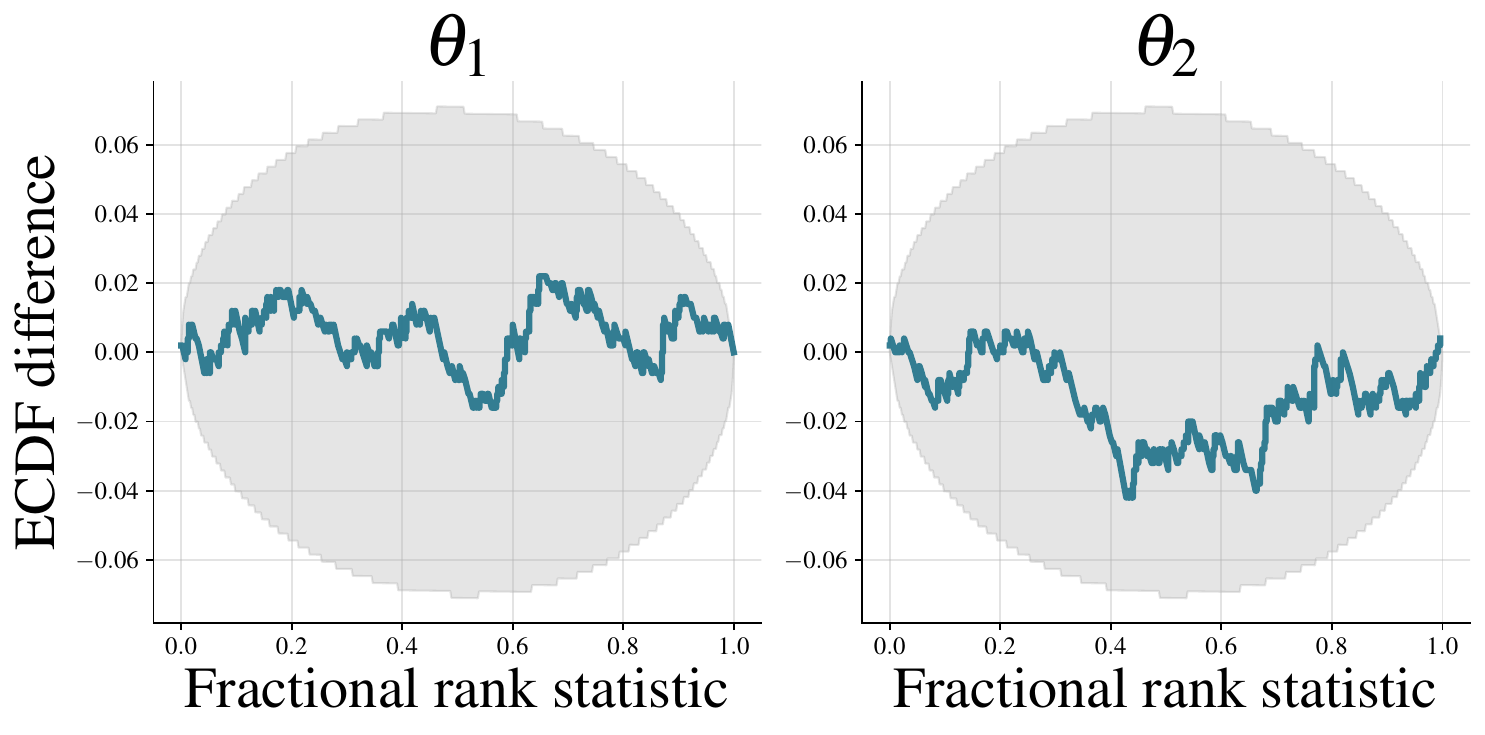}&
         &
         \includegraphics[width=0.40\linewidth]{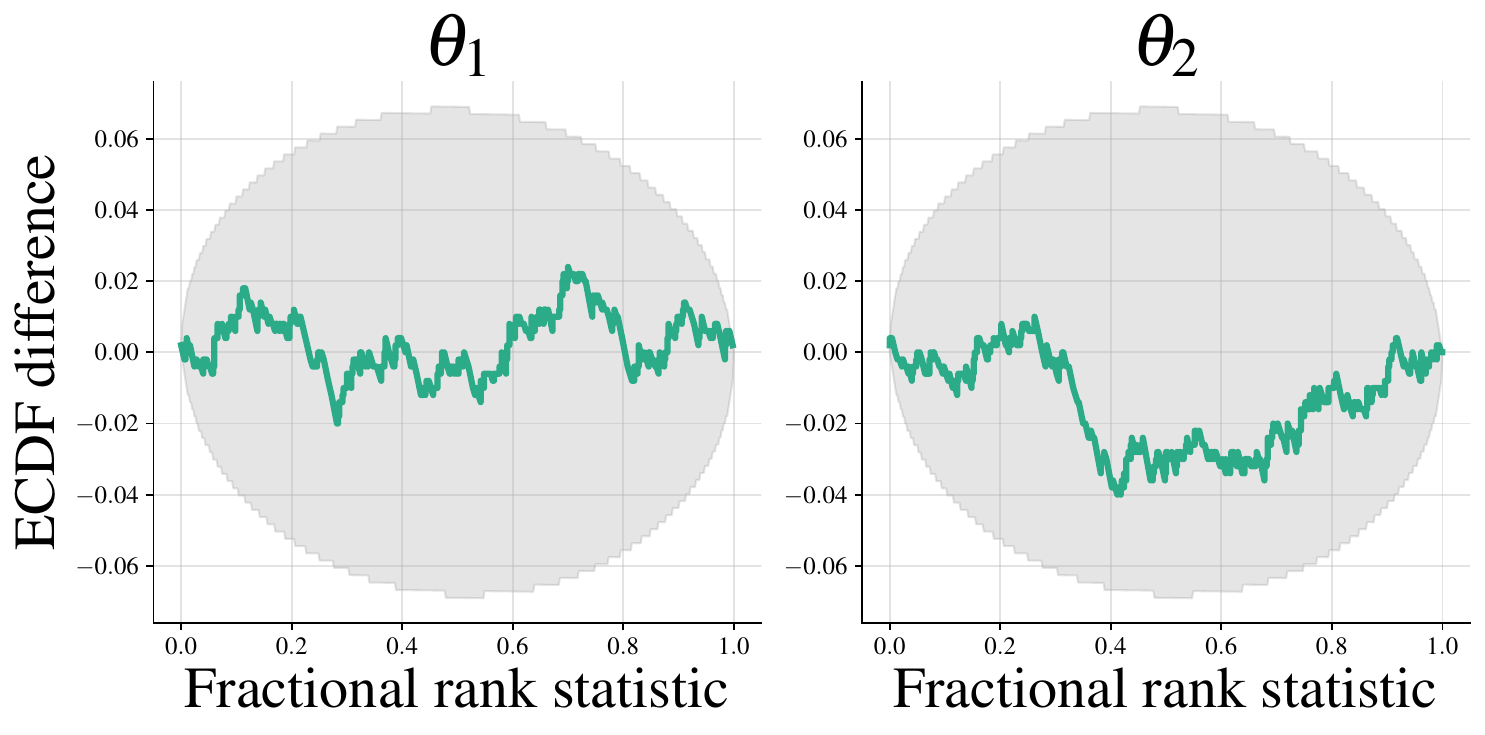}\\ 
    \end{tabular}
    \caption{\textbf{Experiment 1 (Gaussian mixture model).} All approximators are well-calibrated.}\label{fig:app:gmm-npe-sbc}
\end{figure}

All approximators (NPE and self-consistent ones) are well-calibrated according to simulation-based calibration \cite{talts2018validating,sailynoja2022graphical}, as illustrated in \autoref{fig:app:gmm-npe-sbc}.

\subsection{Ablation: Number of Monte Carlo Samples for the Self-Consistency Loss}\label{app:gmm-sc-samples}
In this ablation, we investigate the effect of the number of Monte Carlo samples $K\in\{10, 100, 500\}$ to estimate the variance in Eq.~\ref{eq:loss_npe}.
All self-consistent approximators outperform the NPE baseline with respect via (i) visually better posterior samples and density as well as (ii) a better (lower) MMD on 50 unseen test instances.

\begin{figure*}[H]
    \centering
    \begin{minipage}{0.65\textwidth}
        \begin{subfigure}[t]{\linewidth}
            \textbf{A}
            \includegraphics[width=\linewidth,valign=t]{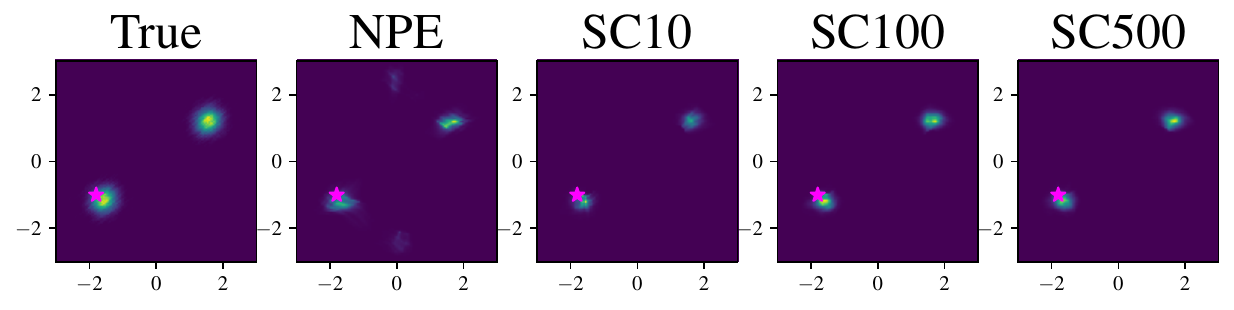}
        \end{subfigure}\\
        \begin{subfigure}[t]{\textwidth}
            \textbf{B}
            \includegraphics[width=\linewidth,valign=t]{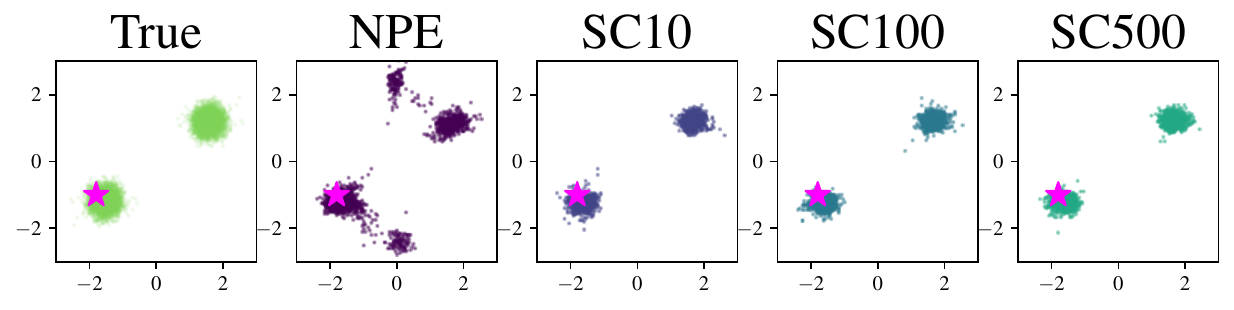}
        \end{subfigure}
    \end{minipage}
    \hfill
    \begin{minipage}{0.32\textwidth}
    \begin{subfigure}[t]{\linewidth}
        \textbf{C}
        \includegraphics[width=\linewidth,valign=t]{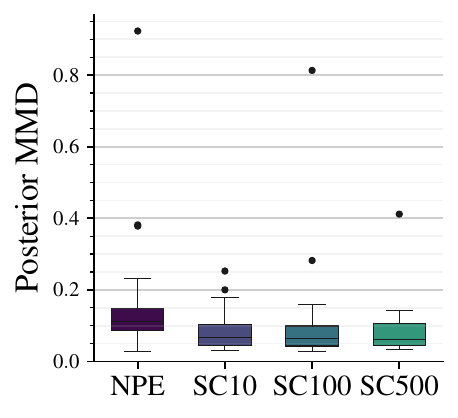}
    \end{subfigure}
    \end{minipage}
    \caption{\textbf{Experiment 1 (Gaussian mixture model).} Our self-consistent estimator (SC10, SC100, SC500) outperforms the NPE baseline using the same neural architecture trained on an identical simulation budget. Adding the self-consistency loss (SC) improves density estimation (\textbf{A}) and sampling (\textbf{B}), judged both visually and via MMD between approximate and true posteriors (\textbf{C}).
    Pink star {\color{magenta}$\boldsymbol{\star}$} marks the true parameter $\thetab$.}%
    \label{fig:gmm-sc-samples}
\end{figure*}

\subsection{Variation: Implicit Likelihood (NPLE)}\label{app:gmm-nple}

We repeat \textbf{Experiment 1} with a fully simulation-based approach which does not need an explicit likelihood to estimate the self-consistency loss.
To this end, we replace the explicit likelihood $p(\Y\given\thetab)$ in Eq.~\ref{eq:sc-loss} with the approximate likelihood $q_{\eta}(\Y\given\thetab)$, which is represented by a neural network and learned simultaneously with the neural posterior approximator.
The full loss follows as
\begin{equation}
\begin{aligned}
    \mathcal{L}_{\text{SC-NPLE}}=\mathbb{E}_{p(\thetab, \Y)}\Big[&
    \underbrace{-\log q_{\phib}(\thetab\given \Y)
    - \log q_{\etab}(\Y\given\thetab)}_{\text{NPLE loss}} \\
    &+
    \underbrace{\lambda\Var_{\thetasc\sim \pi(\thetab)}\Big(
        \log p(\thetasc) + \log q_{\etab}(\Y\given \thetasc) - \log q_{\phib}(\thetasc\given \Y)
    \Big)}_{\text{self-consistency loss $\mathcal{L}_{\text{SC}}$ with approximate likelihood\;}}
    \Big],
\end{aligned}
\end{equation}

which is a combination of the NPLE loss and the self-consistency loss with approximate likelihood.

The simulation budget is fixed to $N=1024$ data sets.
Both NPLE and our self-consistent approximator with $K=100$ Monte Carlo samples are trained for 35 epochs, have an identical neural spline flow architecture, have a heavy-tailed Student-$t_{100}$ latent space \cite{alexanderson2020studentising}, and use an identical DeepSet \cite{zaheer2017deepsets} to learn summary statistics of the data $\Y$ for the posterior approximator.
Since the Monte Carlo approximation in the self-consistency loss now depends on both an approximate posterior and an approximate likelihood, we only activate the self-consistency loss after 20 epochs (as opposed to 5 epochs in \textbf{Experiment 1} with an explicit likelihood).

\begin{figure}[H]
    \centering
    \begin{minipage}{0.38\textwidth}
        \begin{subfigure}[t]{\linewidth}
            \textbf{A}
            \includegraphics[width=\linewidth,valign=t]{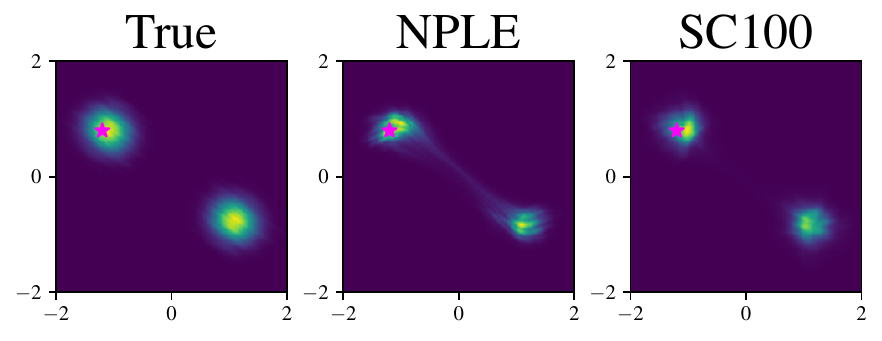}
        \end{subfigure}\\
        \begin{subfigure}[t]{\textwidth}
            \textbf{B}
            \includegraphics[width=\linewidth,valign=t]{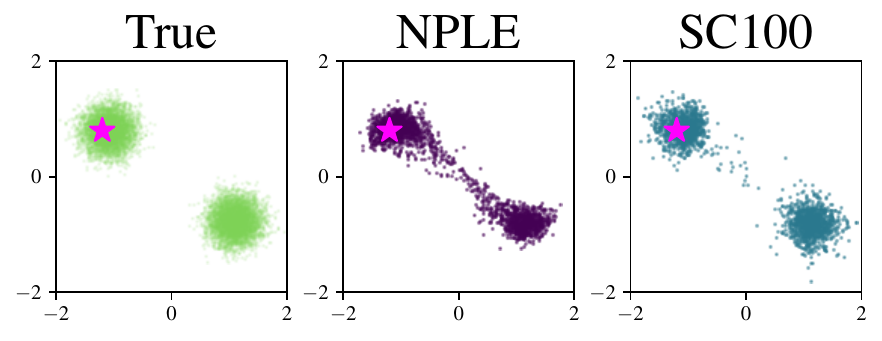}
        \end{subfigure}
    \end{minipage}
    \hfill
    \begin{minipage}{0.28\textwidth}
    \begin{subfigure}[t]{\linewidth}
        \textbf{C}
        \includegraphics[width=\linewidth,valign=t]{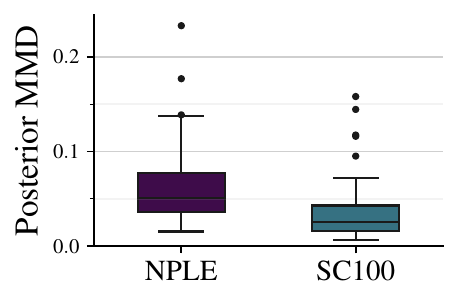}
    \end{subfigure}
    \end{minipage}
    \hfill
    \begin{minipage}{0.28\textwidth}
        \begin{subfigure}[t]{\linewidth}
            \textbf{D}
            \includegraphics[width=\linewidth,valign=t]{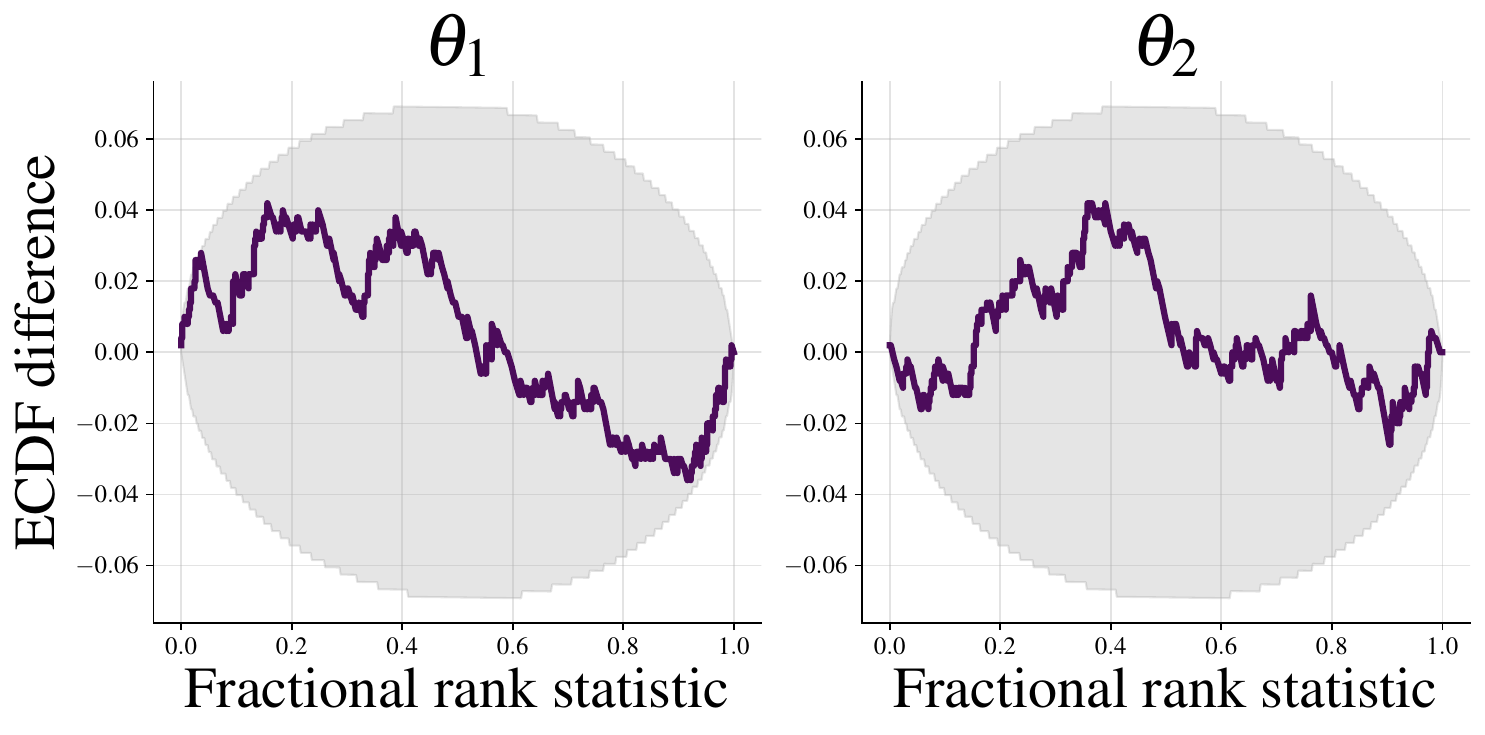}
        \end{subfigure}\\
        \begin{subfigure}[t]{\textwidth}
            \textbf{E}
            \includegraphics[width=\linewidth,valign=t]{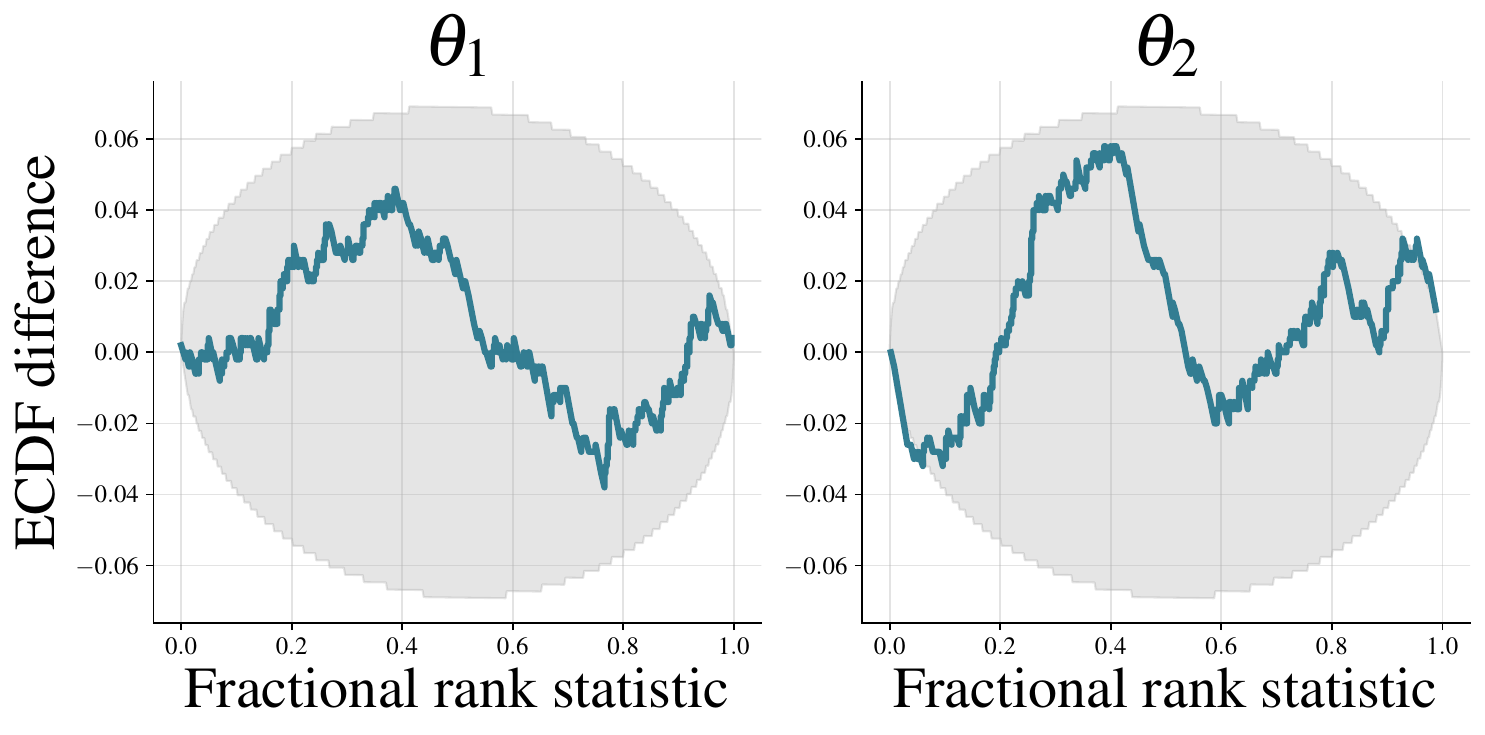}
        \end{subfigure}
    \end{minipage}
    \caption{\textbf{Experiment 1 (Gaussian mixture model).} Our self-consistent posterior estimator (SC100) outperforms the NPLE baseline using the same neural architecture trained on an identical simulation budget. Adding the self-consistency loss leads to improved density estimation (\textbf{A}) and sampling (\textbf{B}), judged both visually and via MMD between approximate and true posteriors (\textbf{C}).
    Both approximators are well-calibrated \textbf{(D, E)}.
    Pink star {\color{magenta}$\boldsymbol{\star}$} marks the true parameter $\thetab$.}
    \label{fig:gmm-nple-figure}
\end{figure}

We can benchmark the methods' performance against the true posterior since the explicit likelihood of this simulator is known (albeit inaccessible for the approximators).
We confirm the results of \textbf{Experiment 1} in the fully simulation-based setting with only an implicit likelihood: Our self-consistent approximator consistently outperforms the baseline NPLE approximator with respect to posterior density and sampling (see \autoref{fig:gmm-nple-figure}).

\subsection{Extension: Sequential Neural Posterior Estimation with Self-Consistency Loss}\label{app:gmm-sc-snpe}

\begin{wrapfigure}{r}{0.45\textwidth}
    \vspace*{-2em}
    \centering
    \includegraphics[width=\linewidth]{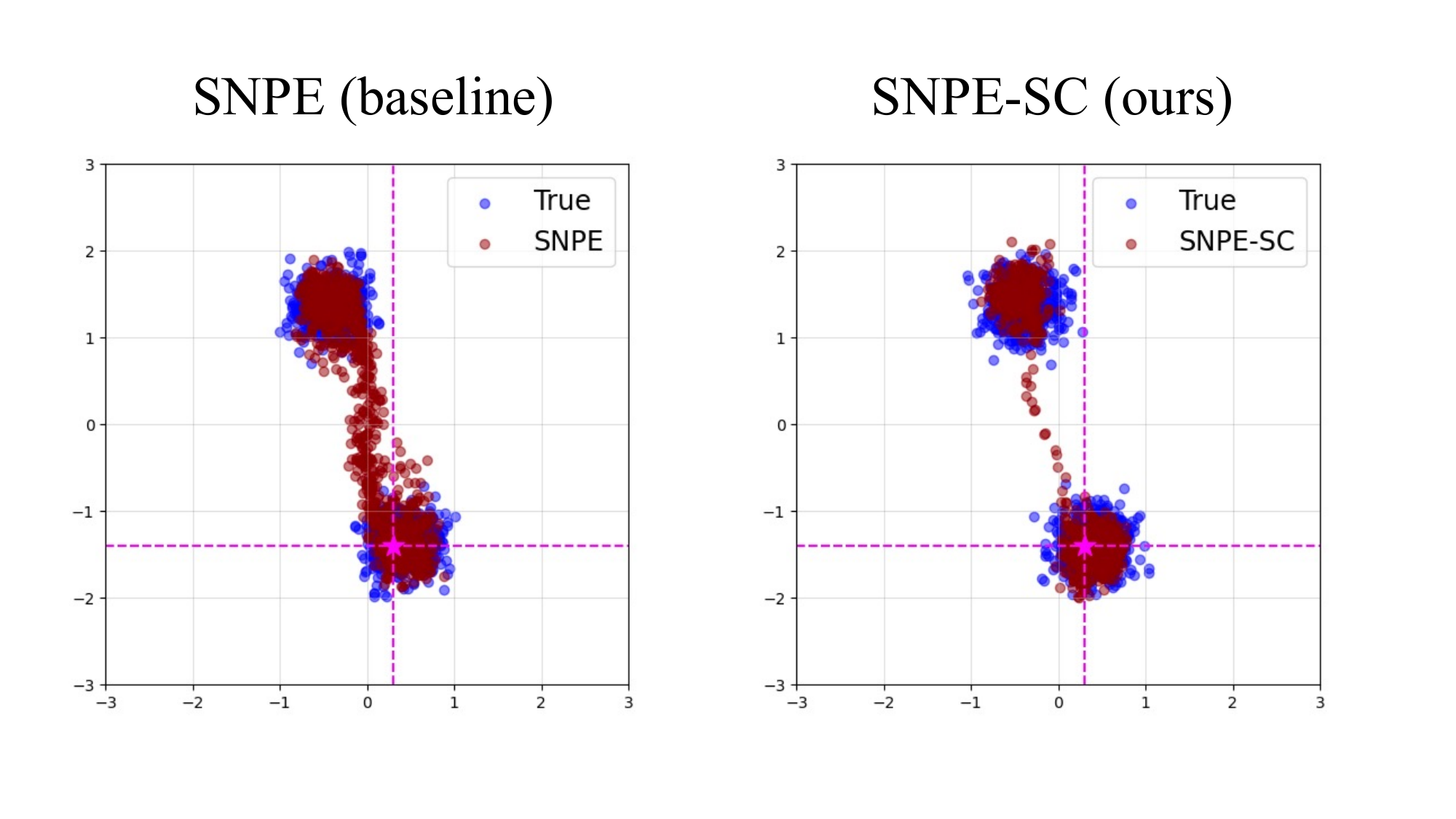}
    \caption{Our self-consistency loss visually improves posterior sampling for SNPE in \textbf{Experiment 1}.}
    \label{fig:snpe-sc}
\end{wrapfigure}

We repeat \textbf{Experiment 1} with sequential neural posterior estimation \citep[SNPE;][]{greenberg2019automatic} and observe similar improvements in the quality of posterior samples by adding our self-consistency loss (see \autoref{fig:snpe-sc}).
This underscores the modular nature of the self-consistency loss, which renders it applicable to a variety of inference algorithms.
We leave an in-depth analysis of the interplay of self-consistency losses and sequential SBI algorithms to future work, as the focus of this paper lies on \emph{amortized} Bayesian inference.

\clearpage
\section{Details about Experiment 2}\label{app:two-moons}
The two moons model is described in detail elsewhere \cite{lueckmann2021benchmarking}.
We use a uniform prior on both components of theta: $\theta_1, \theta_2 \sim \mathrm{Uniform}(-2, 2)$.

The neural network architectures are identical for NPLE (baseline) and SC-NPLE (ours).
The posterior and likelihood networks are identical: Both consist of a neural spline flow \cite{durkan2019neural} with 6 coupling layers of 128 units each and weight regularization with a factor $\gamma=10^{-4}$.
Further, the latent space in the neural spline flow is a heavy-tailed Student-$t$ distribution \cite{alexanderson2020studentising} with 50 degrees of freedom.
The neural networks are trained for 200 epochs with a batch size of 32 and an initial learning rate of $5\cdot10^{-4}$.
For the self-consistent variations, we choose a piecewise constant schedule $\mu(\cdot)$ on the weight $\lambda$, where $\lambda=0$ for the first 100 epochs (i.e., no self-consistency term) and $\lambda=1$ for the remaining 100 epochs.

\section{Details about Experiment 3}\label{app:hes1}

In \textbf{Experiment 3}, we apply our method to an experimental data set in biology \cite{Silk2011}. Upon serum stimulation of various cell lines, the transcription factor Hes1 exhibits sustained oscillatory transcription patterns \cite{Momiji2008}.
The concentration of Hes1 mRNA can be modeled by a set of three differential equations,
\begin{equation}
    \odv{m}{t} = -k_{deg}m + \frac{1}{1+(p_2/p_0)^h},
    \quad
    \odv{p_1}{t} = -k_{deg}p_1 + \nu m - k_1 p_1,
    \quad
    \odv{p_2}{t} = -k_{deg}p_2 + k_1 p_1
\end{equation}
with degradation rate $k_{deg}$, Hes1 mRNA concentration $m$, cytosolic Hes1 protein concentration $p_1$, and nuclear Hes1 protein concentration $p_2$.
The parameters $\thetab=\{p_0, h, k_1, \nu\}$ govern the dynamics of the differential equations, and we estimate them in the unbounded $\log$ space to facilitate inference.
$p_0$ corresponds to the amount of Hes1 protein in the nucleus when the rate of transcription of Hes1 mRNA is at half of its maximum value, $h$ is the Hill coefficient, $k_1$ is the rate of transport of Hes1 protein into the nucleus and $\nu$ is the rate of translation of Hes1 mRNA \cite{Silk2011}.

In accordance to \cite{Filippi2011}, we use fixed initial conditions $m_0=2, p_1=5, p_2=3$ and set $k_{reg}=0.03$.
In our model we regard the observed mRNA concentrations $\y_t$ as noisy measurements of the true underlying mRNA concentration $m_t$ with unit Gaussian observation error, $\y_t \sim \mathcal{N}(m_t, 1)$.

Silk et al.\ \cite{Silk2011} used quantitative real-time PCR to collect the real experimental data
\begin{equation*}
\Y=[1.20, 5.90, 4.58, 2.64, 5.38, 6.42, 5.60, 4.48]
\end{equation*}
where the first observation $\y_1$ is measured after 30 minutes, and all subsequent values are measured in $30$ minute intervals \cite{Filippi2011}. 
The mRNA measures $\y_t$ refer to fold changes relative to a control sample.
The Bayesian model uses Gamma priors on all parameters,

\begin{equation}
    p_0 \sim \Gamma(2, 1),\quad 
    h \sim \Gamma(10, 1),\quad 
    k_1 \sim \Gamma(2, 50),\quad
    \nu \sim \Gamma(2, 50),
\end{equation}
where $\Gamma(a, b)$ denotes the Gamma distribution with shape $a$ and rate $b$.

Both NPLE (baseline) and our self-consistent approximator are trained for 70 epochs and use the same neural spline flow architecture consisting of 4 coupling layers with spectral normalization \cite{miyato2018spectral} and a heavy-tailed Student-$t_{50}$ latent distribution \cite{alexanderson2020studentising} for a more robust latent space. For training, we use a batch size of 16 and an initial learning rate of $10^-3$. For the self-consistent approximator, we use $K=100$ Monte-Carlo samples.  

\begin{figure*}[t]
    \centering
    \begin{subfigure}[t]{0.40\linewidth}
     \stackanchor{\textbf{A}}{}
        \includegraphics[width=\linewidth,valign=t]{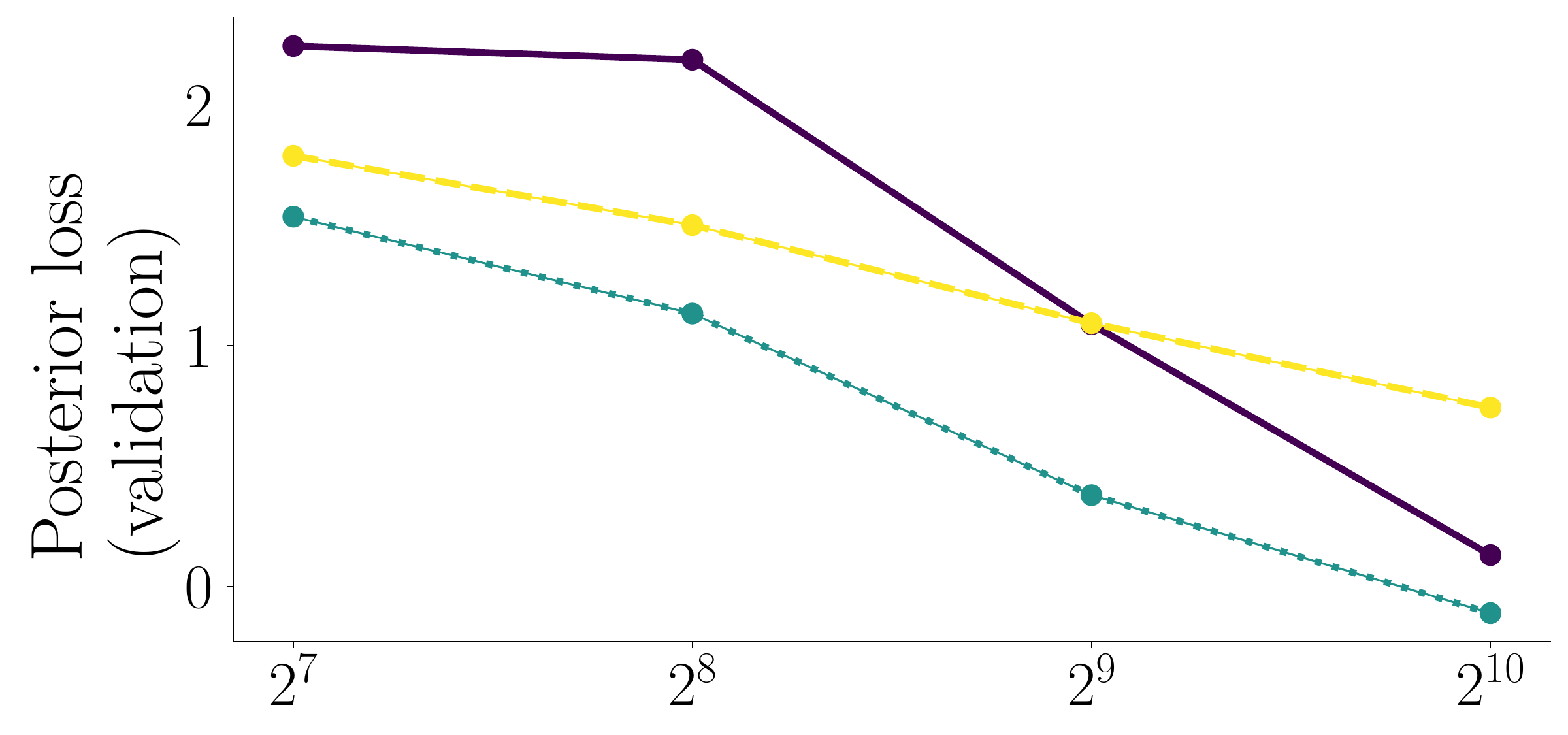}
    \end{subfigure}\hspace*{0.5cm}
    \begin{subfigure}[t]{0.40\linewidth}
     \stackanchor{\textbf{B}}{}
        \includegraphics[width=\linewidth,valign=t]{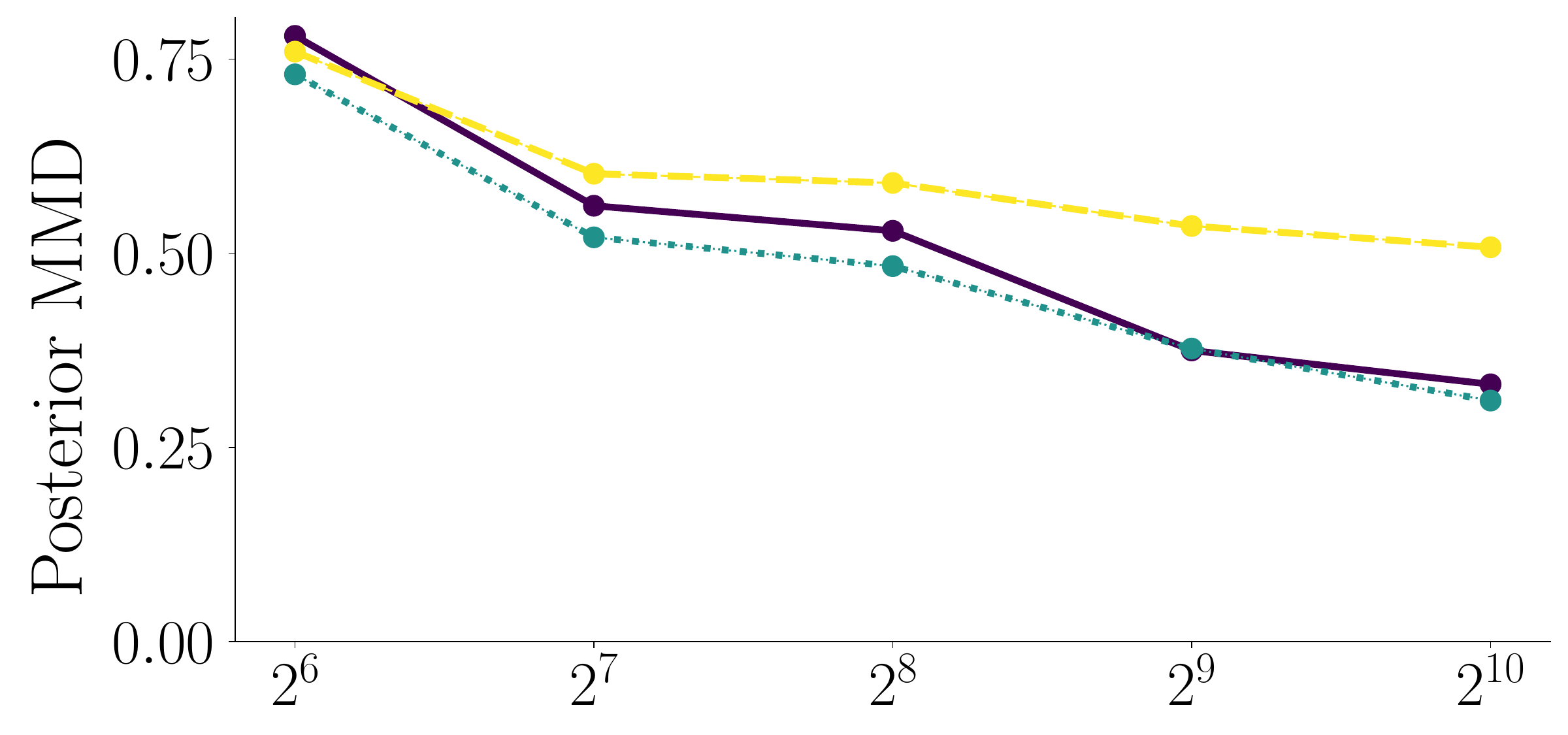}
    \end{subfigure}\hfill
    \\
    \begin{subfigure}[t]{0.80\linewidth}
     \stackanchor{\textbf{C}}{}
        \includegraphics[width=\linewidth,valign=t]{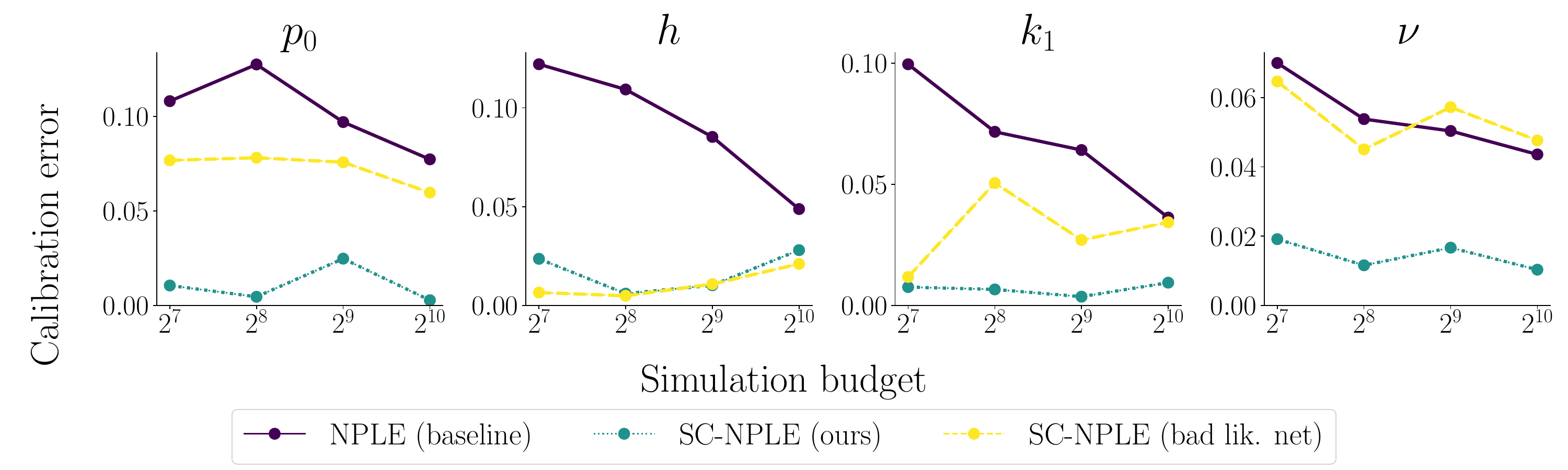}
    \end{subfigure}\hfill
    \caption{\textbf{Experiment 3, Ablation: Underexpressive likelihood network.} For the underexpressive likelihood network, we use a neural spline flow architecture with a single coupling layer, no L2 regularization, no dropout and a linear activation function.
    For low simulation budgets, posterior loss on a separate validation dataset is lower than standard NPLE \textbf{(A)}. However, MMD between the approximate and true posterior is larger for SC-NPLE using the underexpressive likelihood network (\textbf{B}).
    The calibration error (\textbf{C}) lies between the calibration errors obtained from NPLE and SC-NPLE with a more suited likelihood network.}
    \label{fig:hes1-ablation}
\end{figure*}

\begin{figure}[t]
    \centering
    \includegraphics[width=0.9\linewidth]{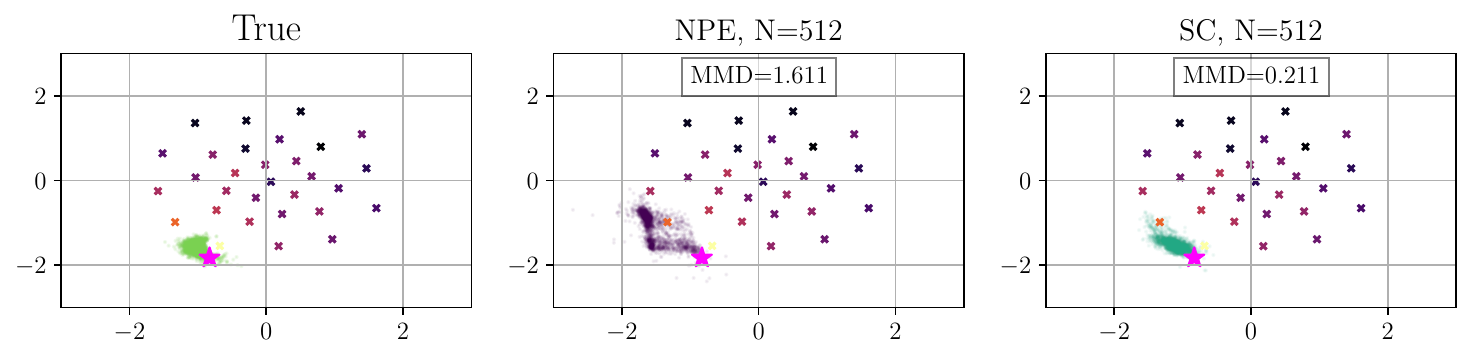}
    \caption{\textbf{Experiment 4 (Source Location Finding).} For one fixed data set, we show the reference posterior (HMC via Stan), as well as NPE (baseline) and SC-NPE (ours). 
    Compared to the NPE baseline, our self-consistent approximator yields a noticeably sharper posterior without introducing additional bias, which is supported by a lower (better) MMD to the reference posterior.}
    \label{fig:enter-label}
\end{figure}

\newpage
\section{Details about Experiment 4}\label{app:source-location}
\begin{wrapfigure}{r}{0.45\linewidth}
    \centering\vspace*{-0.5cm}
    \includegraphics[width=\linewidth]{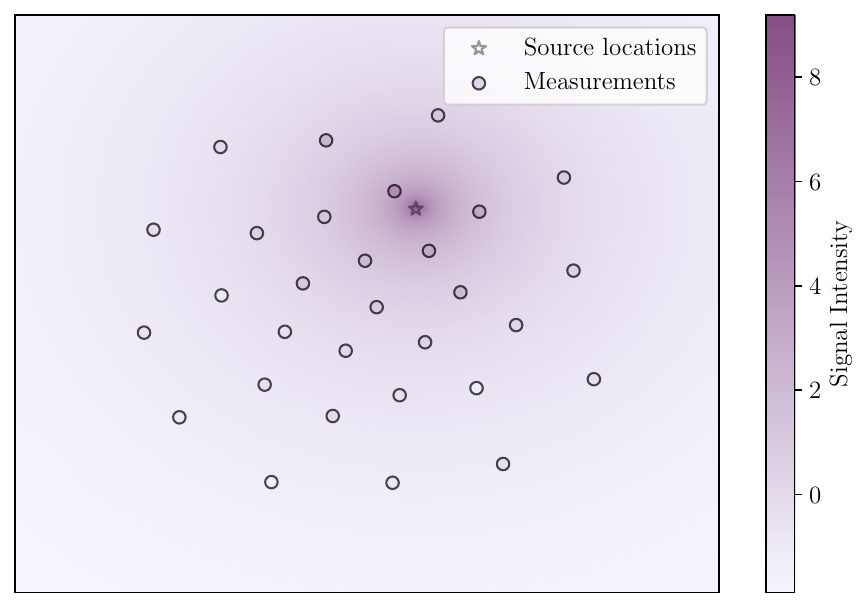}
    \caption{\textbf{Experiment 4 (Source Location Finding).} Illustration of the problem setup.}
    \label{fig:source-location-illustration}
\end{wrapfigure}

The inference task in this experiment is to locate a hidden source $\thetab\in\mathbb{R}^2$ from $N$ noisy measurements $\Y\in\mathbb{R}^{N\times 1}$ of its signal intensity, which is observed at $N$ pre-determined measurement points $\X\in\mathbb{R^{N\time3}}$ (see \autoref{fig:source-location-illustration}). 

More concretely, the sources is sampled from a standard Gaussian: $\thetab  \overset{\text{i.i.d.}}{\sim} \mathcal{N}(\thetab\given\mathbf{0}, \mathbf{I})$ and the likelihood of the outcome is $\Y \sim \mathcal{N}(\Y\given \nu(\thetab, \x), \sigma^2)$, where 
$\nu(\thetab, \X) = b +  \frac{\alpha}{\left( m +  \lvert\lvert \thetab - \X \rvert\rvert \right)^2}$.
For convenience, we concatenate the measurements $\Y$ and measurement point locations $\X$ to a matrix of observables with $(N\times3)$ elements.
In the given context, $\alpha$ may be either predetermined constants or random variables, $b > 0$ represents a fixed background signal, and $m$ is a constant representing the maximum signal. In our experiment we use $\sigma = 0.5, \alpha=1, b=0.1$ and $m=10^{-4}$.

Both baseline NPE and our self-consistent approximator (SC-NPE) use identical neural networks and hyperparameters to ensure a fair comparison.
We use an attention-based permutation-invariant neural network, i.e., a set transformer \cite{lee2019settransformer}, to learn 32-dimensional embeddings that are maximally informative for posterior inference \cite{radev2020bayesflow}.
The posterior network $q_{\phib}$ is a neural spline flow \cite{durkan2019neural} with 6 coupling layers.
The neural networks are trained for 35 epochs with an initial learning rate of $10^{-3}$ and a batch size of 32.

\section{Details about Experiment 5}\label{app:sir_model}

The experiment setup follows \citet{lueckmann2021benchmarking} (Section T.9), with a modification:  we use 160 instead of 10 evenly spaced time points. 
The neural network architectures are identical for NPLE (baseline) and SC-NPLE (ours). 
The posterior and likelihood networks also share same architectures: we use neural spline flow \citep{durkan2019neural} with 5 coupling layers. 
The latent space of the flow is a Student-$t$ distribution with 30 degrees of freedom.
We train for 100 epochs, batch size of 32 and initial learning rate of  $10^{-3}$, simultaneously learning the amortized likelihood and posterior. 
For the self-consistent variations, we apply a piecewise
constant schedule to the weight $\lambda$: $\lambda = 0$ for the first 2 epochs (i.e., no self-consistency term) and $\lambda=0.1$ for
the remaining epochs.

\end{document}